\ifdefined\XeTeXversion\else
  \pdfoutput=1
\fi
\PassOptionsToPackage{table}{xcolor}
\documentclass[11pt,letterpaper,shortlabels]{berkeley}

\usepackage{xurl}
\usepackage{hyperref}
\usepackage[capitalize,noabbrev]{cleveref}
\usepackage[authoryear,round]{natbib}
\usepackage[all]{hypcap}

\usepackage{amsmath,amssymb,mathtools,amsthm,mathrsfs}
\usepackage{nicefrac,dsfont}

\usepackage{algorithm,algorithmic}
\usepackage{multirow,makecell,caption,subcaption,tabularx}
\usepackage{graphicx,booktabs,array}
\usepackage[section]{placeins}
\usepackage{microtype,xspace,enumitem,setspace,soul}
\usepackage{xcolor}
\usepackage{tikz}
\usepackage{pgfplots}
\pgfplotsset{compat=1.18}
\usetikzlibrary{arrows.meta,positioning,fit,calc,shapes.geometric}

\definecolor{AWblue}{RGB}{24,73,117}
\definecolor{AWteal}{RGB}{20,126,132}
\definecolor{AWgold}{RGB}{189,137,43}
\definecolor{AWred}{RGB}{171,55,55}
\definecolor{AWgray}{RGB}{245,247,249}
\definecolor{AWnavy}{RGB}{28,54,82}
\definecolor{AWcyan}{RGB}{30,116,128}
\definecolor{AWsage}{RGB}{65,125,108}
\definecolor{AWamber}{RGB}{176,117,34}
\definecolor{AWcoral}{RGB}{166,72,74}
\definecolor{AWviolet}{RGB}{99,86,139}
\definecolor{AWslate}{RGB}{91,106,121}
\definecolor{AWline}{RGB}{210,219,227}
\hypersetup{
  pdftitle={AlgoWorlds: Benchmarking Tool Use for Global Optimization in Algorithmic Worlds},
  pdfauthor={Zixiang Xu, Jiaan Wang, Fandong Meng},
  colorlinks=true,
  linkcolor=AWblue,
  citecolor=AWteal,
  urlcolor=AWblue,
  filecolor=AWblue
}

\Crefformat{equation}{#2Eq.\;(#1)#3}
\Crefformat{figure}{#2Figure #1#3}
\Crefformat{table}{#2Table #1#3}
\Crefformat{section}{#2Section #1#3}
\Crefname{appendix}{Appendix}{Appendices}

\newcommand{\algoworlds}{\textsc{AlgoWorlds}\xspace}

\newcommand{\awtablestandard}{%
  \small
  \setlength{\tabcolsep}{3pt}%
  \renewcommand{\arraystretch}{1.12}%
  \arrayrulecolor{black}%
}
\newcommand{\awtablecompact}{%
  \footnotesize
  \setlength{\tabcolsep}{2.5pt}%
  \renewcommand{\arraystretch}{1.12}%
  \arrayrulecolor{black}%
}
\newcommand{\awtablezebra}[1]{\rowcolors{#1}{AWgray}{white}}
\newcommand{\awtableheader}{\rowcolor{AWblue!12}}

\newcommand{\ResultBestExact}{38.61\%\xspace}

\newcommand{\ResultSolComplete}{89.86\%\xspace}

\newcommand{\ResultOpusExact}{38.61\%\xspace}

\newcommand{\ResultSolCompleteNonExactRate}{58.42\%\xspace}

\author[1,2,\textdagger]{Zixiang Xu}
\author[1,*]{Jiaan Wang}
\author[1,*]{Fandong Meng}
\affil[1]{Weixin AI, Tencent}
\affil[2]{University of Southern California}
\correspondingauthor={Jiaan Wang (\href{mailto:torchwang@tencent.com}{\nolinkurl{torchwang@tencent.com}});
Fandong Meng (\href{mailto:fandongmeng@tencent.com}{\nolinkurl{fandongmeng@tencent.com}})\\
\textsuperscript{\textdagger}Work was done when Zixiang Xu was interning at Weixin AI, Tencent Inc, China.}

\title{AlgoWorlds: Benchmarking Tool Use for Global Optimization in Algorithmic Worlds}

\begin{abstract}
Tool-use benchmarks generally evaluate whether an agent completes a workflow using appropriate tools and valid arguments.
However, feasibility alone is insufficient in real-world decision settings such as route planning and fleet dispatch. Individual choices interact through shared constraints and costs, so a feasible solution may still be substantially suboptimal. This raises a harder question: can an agent turn information gathered through tools into a globally optimal decision?
In this paper, we introduce \algoworlds, a benchmark that transforms formally specified combinatorial optimization problems into partially observed decision environments with global optima. Each environment contains a hidden optimization instance that the agent observes only through sequential calls to task-specific information tools. The agent then commits to one structured decision, which an independent checker evaluates for
feasibility, objective value, and optimality.
\algoworlds contains 240 such environments, covering ten combinatorial optimization families and four workload levels.
To make these environments both controllable and auditable, family-specific deterministic programs generate hidden instances, while exact algorithms certify their global optima and determine their workload levels. Each instance is exposed through two structurally different multi-tool interfaces, which preserve the same decision problem under different information presentation.
We evaluate seven leading LLMs, including Claude Opus 4.8 and GPT-5.6 Sol, on \algoworlds. Results show that achieving global optimality remains highly challenging: although leading models achieve feasibility in most cases, the best-performing model achieves exact optimality in only \ResultBestExact of cases.
Further analysis shows that even when the LLM agents collect sufficient information to reconstruct the hidden instance, most failures end in feasible but suboptimal decisions. The challenge therefore extends beyond information acquisition to information integration, global constraint reasoning, and decision verification.
\par\smallskip
Homepage: \url{https://xzx34.github.io/AlgoWorlds/}
\par
Code: \url{https://github.com/xzx34/AlgoWorlds}
\end{abstract}

\begin{document}
\maketitle

\section{Introduction}
\label{sec:introduction}

Tool-use benchmarks evaluate whether an agent selects appropriate tools and completes a prescribed workflow \citep{li2023apibank,yan2024bfcl,lu2025toolsandbox,yao2025taubench}.
These benchmarks generally assess correct tool use and task completion, but neglect whether the solution is globally optimal.
In many real-world decision problems, such as route planning, fleet dispatch, and resource allocation, a feasible solution may still be far from optimal \citep{xie2024travelplanner,zhang2026deepplanning,qin2025compass}.
The difficulty is that individual decisions interact through shared constraints and aggregate costs.
For example, in the fleet dispatch problem, a dispatcher must assign every job to a vehicle using job requirements, vehicle eligibility, capacities, route costs, and global activation costs distributed across separate information sources.
Assigning each job to its cheapest eligible route can violate capacity constraints, while a capacity-feasible assignment can remain substantially suboptimal if it activates an unnecessarily costly set of vehicles.
Solving such problems therefore requires multi-source information integration and joint reasoning rather than considering each tool use independently.
We therefore ask: \emph{Can an LLM agent make a globally optimal decision?}

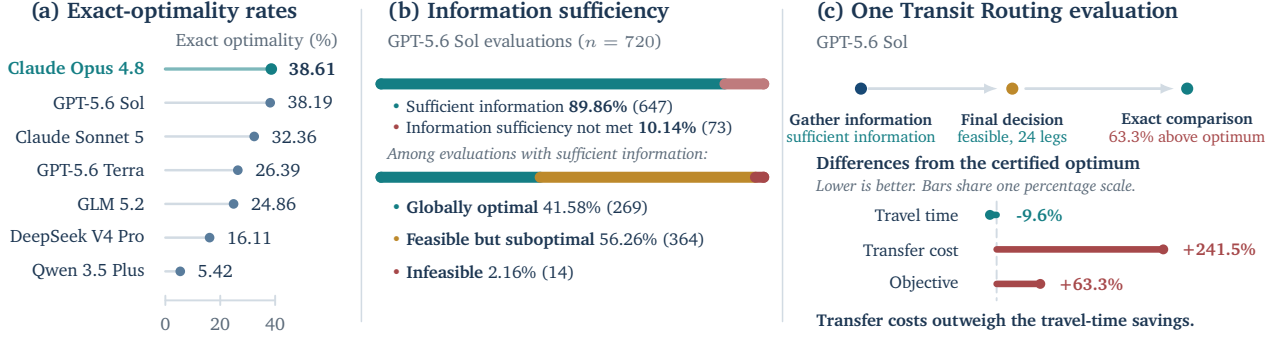
\begin{figure}[t]
  \centering
  \resizebox{\linewidth}{!}{\begin{tikzpicture}[font=\sffamily,line cap=round,line join=round]
\path[use as bounding box] (-0.05,-3.74) -- (15.85,-3.74) -- (15.85,0.62) -- (-0.05,0.62) -- cycle;
\node[anchor=west,font=\bfseries\fontsize{8.2}{9.2}\selectfont,text=AWnavy] at (0.00,0.43) {(a) Exact-optimality rates};
\node[anchor=west,font=\fontsize{6.5}{7.2}\selectfont,text=AWslate] at (1.85,0.05) {Exact optimality (\%)};
\node[anchor=east,font=\bfseries\fontsize{6.5}{7.2}\selectfont,text=AWteal] at (1.72,-0.300) {Claude Opus 4.8};
\draw[line width=1.2pt,AWteal!55] (1.850,-0.300) -- (3.201,-0.300);
\fill[AWteal] (3.201,-0.300) circle (2.1pt);
\node[anchor=west,font=\fontsize{6.5}{7.2}\selectfont,text=AWnavy] at (3.291,-0.300) {\textbf{38.61}};
\node[anchor=east,font=\fontsize{6.5}{7.2}\selectfont,text=AWnavy] at (1.72,-0.730) {GPT-5.6 Sol};
\draw[line width=0.9pt,AWline] (1.850,-0.730) -- (3.187,-0.730);
\fill[AWblue!72] (3.187,-0.730) circle (1.75pt);
\node[anchor=west,font=\fontsize{6.5}{7.2}\selectfont,text=AWnavy] at (3.277,-0.730) {38.19};
\node[anchor=east,font=\fontsize{6.5}{7.2}\selectfont,text=AWnavy] at (1.72,-1.160) {Claude Sonnet 5};
\draw[line width=0.9pt,AWline] (1.850,-1.160) -- (2.983,-1.160);
\fill[AWblue!72] (2.983,-1.160) circle (1.75pt);
\node[anchor=west,font=\fontsize{6.5}{7.2}\selectfont,text=AWnavy] at (3.073,-1.160) {32.36};
\node[anchor=east,font=\fontsize{6.5}{7.2}\selectfont,text=AWnavy] at (1.72,-1.590) {GPT-5.6 Terra};
\draw[line width=0.9pt,AWline] (1.850,-1.590) -- (2.774,-1.590);
\fill[AWblue!72] (2.774,-1.590) circle (1.75pt);
\node[anchor=west,font=\fontsize{6.5}{7.2}\selectfont,text=AWnavy] at (2.864,-1.590) {26.39};
\node[anchor=east,font=\fontsize{6.5}{7.2}\selectfont,text=AWnavy] at (1.72,-2.020) {GLM 5.2};
\draw[line width=0.9pt,AWline] (1.850,-2.020) -- (2.720,-2.020);
\fill[AWblue!72] (2.720,-2.020) circle (1.75pt);
\node[anchor=west,font=\fontsize{6.5}{7.2}\selectfont,text=AWnavy] at (2.810,-2.020) {24.86};
\node[anchor=east,font=\fontsize{6.5}{7.2}\selectfont,text=AWnavy] at (1.72,-2.450) {DeepSeek V4 Pro};
\draw[line width=0.9pt,AWline] (1.850,-2.450) -- (2.414,-2.450);
\fill[AWblue!72] (2.414,-2.450) circle (1.75pt);
\node[anchor=west,font=\fontsize{6.5}{7.2}\selectfont,text=AWnavy] at (2.504,-2.450) {16.11};
\node[anchor=east,font=\fontsize{6.5}{7.2}\selectfont,text=AWnavy] at (1.72,-2.880) {Qwen 3.5 Plus};
\draw[line width=0.9pt,AWline] (1.850,-2.880) -- (2.040,-2.880);
\fill[AWblue!72] (2.040,-2.880) circle (1.75pt);
\node[anchor=west,font=\fontsize{6.5}{7.2}\selectfont,text=AWnavy] at (2.130,-2.880) {5.42};
\draw[AWline,line width=0.7pt] (1.850,-3.25) -- (3.250,-3.25);
\draw[AWline,line width=0.7pt] (1.850,-3.19) -- (1.850,-3.31);
\node[anchor=north,font=\fontsize{6.5}{7.2}\selectfont,text=AWslate] at (1.850,-3.34) {0};
\draw[AWline,line width=0.7pt] (2.550,-3.19) -- (2.550,-3.31);
\node[anchor=north,font=\fontsize{6.5}{7.2}\selectfont,text=AWslate] at (2.550,-3.34) {20};
\draw[AWline,line width=0.7pt] (3.250,-3.19) -- (3.250,-3.31);
\node[anchor=north,font=\fontsize{6.5}{7.2}\selectfont,text=AWslate] at (3.250,-3.34) {40};
\draw[AWline,line width=0.55pt] (4.35,0.46) -- (4.35,-3.52);
\node[anchor=west,font=\bfseries\fontsize{8.2}{9.2}\selectfont,text=AWnavy] at (4.550,0.43) {(b) Information sufficiency};
\node[anchor=west,font=\fontsize{6.5}{7.2}\selectfont,text=AWslate] at (4.550,0.05) {GPT-5.6 Sol evaluations ($n=720$)};
\draw[AWteal,line width=4.0pt] (4.600,-0.49) -- (8.985,-0.49);
\draw[AWcoral!72,line width=4.0pt] (8.985,-0.49) -- (9.480,-0.49);
\fill[AWteal] (4.600,-0.49) circle (2.0pt);
\fill[AWcoral!72] (9.480,-0.49) circle (2.0pt);
\node[anchor=west,font=\fontsize{6.05}{6.8}\selectfont,text=AWnavy] at (4.60,-0.78) {\textcolor{AWteal}{\textbullet} Sufficient information \textbf{89.86\%} (647)};
\node[anchor=west,font=\fontsize{6.05}{6.8}\selectfont,text=AWnavy] at (4.60,-1.06) {\textcolor{AWcoral}{\textbullet} Information sufficiency not met \textbf{10.14\%} (73)};
\node[anchor=west,font=\itshape\fontsize{5.9}{6.6}\selectfont,text=AWslate] at (4.550,-1.39) {Among evaluations with sufficient information:};
\draw[AWteal,line width=4.0pt] (4.600,-1.68) -- (6.629,-1.68);
\draw[AWgold,line width=4.0pt] (6.629,-1.68) -- (9.374,-1.68);
\draw[AWcoral,line width=4.0pt] (9.374,-1.68) -- (9.480,-1.68);
\fill[AWteal] (4.600,-1.68) circle (2.0pt);
\fill[AWcoral] (9.480,-1.68) circle (2.0pt);
\node[anchor=west,font=\fontsize{6.25}{7.0}\selectfont,text=AWnavy] at (4.60,-2.08) {\textcolor{AWteal}{\textbullet} \textbf{Globally optimal} 41.58\% (269)};
\node[anchor=west,font=\fontsize{6.25}{7.0}\selectfont,text=AWnavy] at (4.60,-2.49) {\textcolor{AWgold}{\textbullet} \textbf{Feasible but suboptimal} 56.26\% (364)};
\node[anchor=west,font=\fontsize{6.25}{7.0}\selectfont,text=AWnavy] at (4.60,-2.90) {\textcolor{AWcoral}{\textbullet} \textbf{Infeasible} 2.16\% (14)};
\draw[AWline,line width=0.55pt] (9.72,0.46) -- (9.72,-3.52);
\node[anchor=west,font=\bfseries\fontsize{8.2}{9.2}\selectfont,text=AWnavy] at (10.02,0.43) {(c) One Transit Routing evaluation};
\node[anchor=west,font=\fontsize{6.5}{7.2}\selectfont,text=AWslate] at (10.02,0.05) {GPT-5.6 Sol};
\draw[-{Latex[length=1.8mm]},AWline,line width=1.0pt] (10.82,-0.55) -- (12.47,-0.55);
\draw[-{Latex[length=1.8mm]},AWline,line width=1.0pt] (12.82,-0.55) -- (14.72,-0.55);
\fill[AWblue] (10.72,-0.55) circle (2.2pt);
\fill[AWgold] (12.65,-0.55) circle (2.2pt);
\fill[AWteal] (14.88,-0.55) circle (2.2pt);
\node[anchor=north,align=center,font=\fontsize{5.8}{6.5}\selectfont,text=AWnavy] at (10.72,-0.72) {\textbf{Gather information}\\\textcolor{AWteal}{sufficient information}};
\node[anchor=north,align=center,font=\fontsize{5.8}{6.5}\selectfont,text=AWnavy] at (12.65,-0.72) {\textbf{Final decision}\\\textcolor{AWteal}{feasible, 24 legs}};
\node[anchor=north,align=center,font=\fontsize{5.8}{6.5}\selectfont,text=AWnavy] at (14.88,-0.72) {\textbf{Exact comparison}\\\textcolor{AWcoral}{63.3\% above optimum}};
\node[anchor=west,font=\bfseries\fontsize{6.5}{7.2}\selectfont,text=AWnavy] at (10.02,-1.52) {Differences from the certified optimum};
\node[anchor=west,font=\itshape\fontsize{5.8}{6.5}\selectfont,text=AWslate] at (10.02,-1.81) {Lower is better. Bars share one percentage scale.};
\draw[AWline,densely dashed,line width=0.7pt] (12.45,-1.99) -- (12.45,-3.25);
\node[anchor=east,font=\fontsize{6.2}{7.0}\selectfont,text=AWnavy] at (12.10,-2.16) {Travel time};
\draw[AWteal,line width=2.4pt] (12.365,-2.16) -- (12.45,-2.16);
\fill[AWteal] (12.365,-2.16) circle (1.8pt);
\node[anchor=west,font=\bfseries\fontsize{6.2}{7.0}\selectfont,text=AWteal] at (12.57,-2.16) {-9.6\%};
\node[anchor=east,font=\fontsize{6.2}{7.0}\selectfont,text=AWnavy] at (12.10,-2.60) {Transfer cost};
\draw[AWcoral,line width=2.4pt] (12.45,-2.60) -- (14.575,-2.60);
\fill[AWcoral] (14.575,-2.60) circle (1.8pt);
\node[anchor=west,font=\bfseries\fontsize{6.2}{7.0}\selectfont,text=AWcoral] at (14.675,-2.60) {+241.5\%};
\node[anchor=east,font=\fontsize{6.2}{7.0}\selectfont,text=AWnavy] at (12.10,-3.04) {Objective};
\draw[AWcoral,line width=2.4pt] (12.45,-3.04) -- (13.007,-3.04);
\fill[AWcoral] (13.007,-3.04) circle (1.8pt);
\node[anchor=west,font=\bfseries\fontsize{6.2}{7.0}\selectfont,text=AWcoral] at (13.107,-3.04) {+63.3\%};
\node[anchor=west,font=\bfseries\fontsize{6.2}{7.0}\selectfont,text=AWnavy] at (10.02,-3.52) {Transfer costs outweigh the travel-time savings.};
\end{tikzpicture}}
    \caption{\textbf{Sufficient information does not guarantee global optimality.} \textbf{(a)} Exact-optimality rates for the seven evaluated LLMs. \textbf{(b)} In the evaluation of GPT-5.6 Sol, trajectories are initially categorized based on whether they gather sufficient information. Those that meet this criterion are further classified according to their final-decision results. \textbf{(c)} An example evaluation of GPT-5.6 Sol on Transit Routing. The trajectory gathers sufficient information and ends with a feasible, albeit non-optimal, route.}
  \label{fig:results-teaser}
\end{figure}

We introduce \algoworlds, a benchmark for evaluating whether LLM agents can turn information acquired through tools into globally optimal decisions.
We define an \emph{algorithmic world} as a decision environment comprising a query, a hidden problem instance, and a set of task-specific tools.
Among them, the instance contains the case-specific information that determines the feasible decisions and their objective values, while the tools provide access to different parts of this information.
When addressing the task, an LLM agent receives the query and the tools, but not the instance, and should obtain the information needed to make its decision through tool calls.
To simulate the limitations of information access in real-world scenarios, each tool use incurs a specific cost within a predefined total access budget.
Solving this task thus requires multi-turn interaction, in which the agent decides which tool to query next based on what it has already observed, and finally commits to a decision.

Building such worlds reliably requires hidden instances that can be scaled systematically and whose optima can be verified exactly, together with faithful informational tools.
Following this recipe, we instantiate \algoworlds from ten formally specified combinatorial optimization families, \emph{e.g.}, routing, exact cover, facility selection and constrained sequencing.
Within each family, a human-written deterministic generator creates hidden instances based on specified size and structure configurations, enabling us to systematically vary the problem's scale and the underlying reasoning structure of its decisions.
A tailored exact algorithm then solves each instance to global optimality.
We also count the primitive operations the algorithm actually executes to identify four workload levels.
We next build the informational tools that expose different parts of an instance, such as the travel distance between two cities.
Specifically, we present each hidden instance through two structurally different multi-tool interfaces that expose the same underlying decision information. This paired design lets us assess sensitivity to tool presentation while holding the same optimization problem.
To further ensure data quality, we independently confirm each instance's optimal decision, objective value, and uniqueness, verify that the paired interfaces encode the same hidden instance, and confirm that sufficient information can be acquired within the total access budget.

To evaluate LLM agents in \algoworlds, we report three final-decision metrics and two trajectory-oriented diagnostics.
(a) \emph{Final-decision metrics.} Exact optimality, our primary metric, records whether a feasible decision attains the global optimum. Feasibility records whether the decision satisfies the task constraints. Reference utility assigns a feasible decision a continuous score ranging from $0$ to $1$, with the global optimum at $1$ and a predefined suboptimal reference decision at $0$.
(b) \emph{Trajectory-oriented diagnostics.} Information sufficiency is satisfied when the trajectory collects enough tool-provided information to reconstruct the hidden instance and determine its global optimum. Discovery coverage measures how close the information gathered along the trajectory comes to meeting that threshold.

Using \algoworlds, we evaluate seven leading LLMs: Claude Opus 4.8~\citep{anthropic2026opus48}, Claude Sonnet 5 \citep{anthropic2026sonnet5}, GPT-5.6 Sol, GPT-5.6 Terra \citep{openai2026gpt56}, DeepSeek V4 Pro \citep{deepseek2026v4}, Qwen 3.5 Plus \citep{qwen2026qwen35}, and GLM 5.2 \citep{zai2026glm52}. As \Cref{fig:results-teaser}(a) shows, Claude Opus 4.8 achieves the highest exact-optimality rate, at only \ResultOpusExact, followed closely by GPT-5.6 Sol at 38.19\%.
Notably, \ResultSolComplete of GPT-5.6 Sol's trajectories contain sufficient information, yet \ResultSolCompleteNonExactRate of them still end without a globally optimal decision.
Most of these failures end in feasible but suboptimal decisions, as \Cref{fig:results-teaser}(b) shows.
The Transit Routing example in \Cref{fig:results-teaser}(c) illustrates this pattern.
In this example, the trajectory gathers sufficient information, yet GPT-5.6 Sol constructs a feasible route that has lower travel time than the globally optimal route while incurring substantially higher transfer costs. As a result, its overall objective value is 63.3\% higher.
Together, these results expose a substantial gap between information availability and global optimization: current LLM agents often fail to reach the optimum even when they collect sufficient information.

\textbf{Our main contributions are threefold.}
\textbf{(1)} We introduce \algoworlds, a benchmark for evaluating whether LLM agents can turn information acquired through tools into globally optimal decisions. Its 240 algorithmic worlds span ten optimization families and four workload levels.
\textbf{(2)} To build the benchmark, we provide a verified optimum and faithful information tools for every hidden instance. Paired tool interfaces further measure the sensitivity of information presentation while holding the same underlying decision problem.
\textbf{(3)} Across seven leading LLMs, we identify a clear capability profile: agents are often effective at gathering information and producing feasible decisions, but remain much less reliable at finding the global optimum. This profile is relatively stable across paired interfaces but varies sharply across different optimization families in ways that do not follow classical asymptotic complexity alone.

\section{Related Work}
\label{sec:related}

\paragraph{Tool use and interactive agents.}
Tool-use benchmarks evaluate whether an LLM agent can select an appropriate
tool, supply valid arguments, maintain state across turns, and complete a
task using external systems. API-Bank \citep{li2023apibank} and ToolLLM
\citep{qin2023toolllm} provide broad evaluations of API selection and
argument construction. BFCL \citep{yan2024bfcl} focuses on function-calling
accuracy, while ACEBench \citep{chen2025acebench} extends evaluation across
tools, agents, and domains. Interactive benchmarks extend evaluation from
isolated calls to sustained interaction. ToolSandbox
\citep{lu2025toolsandbox} records stateful execution and intermediate
milestones. $\tau$-bench \citep{yao2025taubench}
and $\tau^2$-bench \citep{barres2025tau2bench} evaluate policy-constrained
agent--user interaction. The Tool Decathlon
\citep{li2025tooldecathlon,wang2025dyflow} covers diverse long-horizon
workflows, while TaskBench
\citep{shen2024taskbench} and MCP-Bench \citep{wang2025mcpbench} study task
decomposition and coordination across tools or servers. These benchmarks
together test whether agents can operate tools and complete workflows.
\algoworlds builds on these capabilities by asking whether an agent can use
information acquired through tools to identify a globally optimal decision.

\paragraph{Partially observed environments and information seeking.}
This line of work studies reasoning and action when task information is not
fully available at the beginning \citep{xu2025socialmaze}. AgentBench
\citep{liu2023agentbench} provides interactive environments for evaluating
general agents, and AgentBoard \citep{ma2024agentboard} provides
trajectory-level analysis of their progress. WebArena \citep{zhou2024webarena},
WorkArena \citep{drouin2024workarena}, OSWorld \citep{xie2024osworld}, and
AppWorld \citep{trivedi2024appworld} ground interaction in executable web,
desktop, or application states.
GAIA \citep{mialon2023gaia}, AssistantBench \citep{yoran2024assistantbench}, BrowseComp \citep{wei2025browsecomp}, WideSearch~\citep{ICLR2026_11195878}, and EvoBrowseComp~\citep{wang2026evobrowsecomp} emphasize retrieval and synthesis across multiple sources. Information
foraging theory \citep{pirolli1999information} provides a complementary account
of how information-seeking behavior responds to the value and cost of
information. InForage
\citep{qian2025inforage} and SCOUT \citep{zhang2026scout} bring this perspective
to search-enhanced reasoning and long-context information acquisition.
In \algoworlds, task-specific information tools expose well-defined facts from
a hidden optimization instance. For each algorithmic world, executable
tool-use plans yield fact sets sufficient to reconstruct that
instance. Trajectory diagnostics measure coverage of these sets, while the
final decision is evaluated against a formal objective and an exact global
optimum rather than a textual answer or workflow state.

\paragraph{Planning with consequential decisions.}
Planning benchmarks evaluate whether an agent can produce actions or decisions
that remain coherent under constraints and have consequences beyond an
individual step. PlanBench \citep{valmeekam2023planbench} and ACPBench
\citep{kokel2025acpbench} express symbolic planning problems in controlled
language. NATURAL PLAN \citep{zheng2024naturalplan} evaluates planning over
records provided in context. TravelPlanner \citep{xie2024travelplanner}
requires agents to gather information and construct feasible itineraries under
real-world constraints. DeepPlanning \citep{zhang2026deepplanning} combines
proactive information acquisition with long, constrained workflows, while the
Agent Planning Benchmark \citep{sun2026apb} tests planning in environments with
irrelevant or malfunctioning tools. COMPASS \citep{qin2025compass} evaluates
both feasibility and utility after conversational or tool-mediated information
gathering. These works show that successful planning requires more than valid
individual actions. \algoworlds focuses specifically on globally optimal
decision making: after gathering information through tools, the agent
provides a structured decision whose feasibility, objective value, and exact
global optimality can be evaluated.

\paragraph{Algorithmic and optimization reasoning.}
Work on algorithmic and optimization reasoning studies whether language models
can execute algorithms, formulate optimization problems, or construct
procedures for solving them \citep{xu2026gta}.
CLRS-Text \citep{markeeva2024clrstext} evaluates textual execution of classical
algorithms under controlled input scaling. NL4Opt
\citep{ramamonjison2023nl4opt} and Mamo \citep{huang2024mamo} study the
translation of natural-language descriptions into mathematical optimization
models. OptiBench \citep{yang2024optibench} and ORLM
\citep{huang2024orlm} evaluate automated optimization modeling, while OptMATH
\citep{lu2025optmath} and OptiVerse \citep{zhang2026optiverse} broaden the
coverage of optimization problems and solution processes. CO-Bench
\citep{sun2025cobench} asks agents to generate algorithms for combinatorial
optimization, and HeuriGym \citep{chen2025heurigym} evaluates executable
heuristic construction. In these settings, the problem instance is typically
available before optimization, or the generated solver is itself the target.
\algoworlds instead asks an agent to acquire information about a hidden
instance through task-specific information tools and then provide one
structured decision. Its exact algorithms support benchmark construction and
certify the global optima used for scoring. They are not tools available to the
evaluated agent.

\section{AlgoWorlds}
\label{sec:algoworlds}

\subsection{Formal Definition}
\label{sec:world-definition}

\paragraph{Algorithmic Worlds.}
An \emph{algorithmic world} is a partially observed decision environment containing a textual question ($q$), a hidden optimization instance ($\mathcal{I}$), and task-specific tools ($\mathcal{T}$). To simulate the limitations of information access in real-world scenarios, each tool $t_i\in\mathcal{T}$ incurs a specific cost $c_i$, and there is a predefined budget ($b$) to specify the upper limit of the total cost. Based on the above, we define an algorithmic world as:
\begin{equation} 
  w=\{q, \mathcal{I},\mathcal{T}, f_c, b\}
  \label{eq:world}
\end{equation}
where $f_c$ is the cost function to define the cost of each tool, \emph{i.e.}, $c_i=f_c(t_i)$.
The textual question $q$ describes the decision to be made, the constraints it must satisfy, and the objective to optimize. The instance $\mathcal{I}$ contains the case-specific data that determine the feasible decisions and their objective values.
The task-specific tools $\mathcal{T}$ include various information tools $\mathcal{T}_{\mathrm{info}}$ and one final-decision tool
$t_{\mathrm{dec}}$ for LLM agents to obtain information about $\mathcal{I}$ and commit a final decision, respectively.
The final-decision tool $t_{\mathrm{dec}}$ is zero-cost, and can be invoked only once at the end of one trajectory.
Taking the fleet dispatch problem as an example, $\mathcal{I}$ records job loads, vehicle capacities, eligible vehicle--job pairs, route prices, and fixed activation costs.
The question $q$ asks the agent to assign every job while minimizing route and activation costs subject to eligibility and capacity constraints.
Separate job, route, policy, and vehicle-contract tools in $\mathcal{T}_{\mathrm{info}}$ reveal the information needed to make the optimal assignment. The final-decision tool $t_{\mathrm{dec}}$ accepts one complete job-to-vehicle mapping.

\paragraph{Agent trajectories.}
Let $\pi$ denote an LLM agent. When addressing the task, the agent receives $w\setminus\mathcal{I} = \{q,\mathcal{T}, f_c, b\}$, and interacts with task-specific tools $\mathcal{T}$ to solve the task:
\begin{equation}
\pi(w\setminus\mathcal{I})\rightarrow t_{s_1}\rightarrow o_{s_1} \rightarrow t_{s_2}\rightarrow o_{s_2} \rightarrow \cdots \rightarrow t_{s_k}\rightarrow o_{s_k}\rightarrow t_{\mathrm{dec}} (y),
  \qquad
  \sum_{i=1}^{k} f_c(t_{s_i}) \leq b.
  \label{eq:trajectory}
\end{equation}
where $t_{s_i}$ and $o_{s_i}$ denote the tool call(s) in the $i$-th step and the corresponding returned information (observation). $y$ is the agent's final decision committed via $t_{\mathrm{dec}}$.
At each step, the agent interacts with tools based on the task and the trajectory history.
Information tools can be repeatedly invoked as long as their cumulative cost does not exceed $b$. If a call is rejected due to insufficient remaining budget, it may return an error response; however, it will not be executed, nor will it incur any costs.
\Cref{app:worked-world} presents an example drawn from a released Fleet Dispatch world, illustrating its task, tool interactions, and final decision.

\subsection{Benchmark Construction}
\label{sec:synthesis}

\algoworlds is constructed from ten formally specified combinatorial optimization families. Each family specifies the structure of $\mathcal{I}$, the constraints that make a decision $y$ feasible, and the objective used to evaluate $y$. Although the families share the interaction protocol in \cref{sec:world-definition}, they differ in how individual choices interact under global constraints and in the exact algorithms used to find the global optimum. \Cref{tab:cases} lists all ten families and summarizes their decision, instance data, exact solution method, and number of information tools $\mathcal{T}_{\mathrm{info}}$. Formal problem definitions and complexity analyses are provided in \cref{app:cases}.

\begin{table}[t]
\centering
\begingroup
\awtablecompact
\awtablezebra{2}
\resizebox{0.97\textwidth}{!}{%
\begin{tabular}{@{}
  >{\raggedright\arraybackslash}p{2.7cm}
  >{\raggedright\arraybackslash}p{3.cm}
  >{\raggedright\arraybackslash}p{5.0cm}
  >{\raggedright\arraybackslash}p{4.cm}
  >{\centering\arraybackslash}p{0.9cm}@{}}
\toprule
\awtableheader
Task family & Decision $y$ & Instance data in $\mathcal{I}$ & Exact solution method
& $|\mathcal{T}_{\mathrm{info}}|$ \\
\midrule
Transit Routing &
One route leg per stage &
Stage-specific legs, travel times, line labels, directed transfer costs, and residues &
Residue and last-line dynamic program & 6 \\

Basket Assembly &
A ticket set forming an exact cover &
Requirements, ticket--item incidence, and prices &
First-uncovered-item bitmask dynamic program & 5 \\

Station Siting &
One site per zone &
Shared capacity, site costs, and bounded-span pairwise demand rewards &
Resource-aware band-frontier dynamic program & 4 \\

Authorization Planning &
A package sequence &
Authorization relations, policy adjustments, handoffs, and clearance residue &
Product-state dynamic program & 6 \\

Series Portfolio &
One lot per title &
Lot values, scopes, and overlapping signed factors &
Bounded-span factor dynamic program & 6 \\

Machine Layout &
A machine-to-slot permutation &
Machine flows, a unit-spaced slot order, and placement adjustments &
Cut-identity subset dynamic program & 8 \\

Sequential Matching &
An ordered perfect matching &
Compatibility, pair costs, and directed handoffs &
Used-set and last-match dynamic program & 6 \\

Fleet Dispatch &
A job-to-vehicle assignment &
Job loads, eligible vehicle--job pairs, capacities, route costs, and activation costs &
Mixed-radix capacity dynamic program & 6 \\

Evidence-Joined Routing &
A layered source-to-target path &
Layered graph with source and target, quantities, exceptions, tariffs, and residues &
Expanded-state shortest path & 6 \\

Migration Portfolio &
One package per component group and an activated domain set &
Package costs, support domains, activation costs, and signed interactions &
Activated-domain mask enumeration & 8 \\
\bottomrule
\end{tabular}%
}
\caption{The ten optimization families in \algoworlds.}
\label{tab:cases}
\endgroup
\end{table}

For each family, benchmark construction proceeds in three stages: First, a deterministic generator produces a hidden instance ($\mathcal{I}$) based on a specified size and structure configuration. Second, a tailored exact algorithm solves the instance to find its optimal decision and determines its workload level. Finally, we construct a textual question ($q$) and expose the same instance through Direct and Mediated tool interfaces ($\mathcal{T}_{\mathrm{d}}$ and $\mathcal{T}_{\mathrm{m}}$) with the same call costs ($f_c$) and access budget ($b$), yielding a pair of algorithmic worlds, \emph{i.e.}, $\{q, \mathcal{I},\mathcal{T}_{\mathrm{d}}, f_c, b\}$ and $\{q, \mathcal{I},\mathcal{T}_{\mathrm{m}}, f_c, b\}$.

\paragraph{Stage 1: deterministic instance generation.}
Each family uses a human-written deterministic generator to produce $\mathcal{I}$ based on a specified size and structure configuration.
Among them, the size configuration dictates the scale of the problem, while the structure configuration governs how the decision $y$ interacts within the global constraints and objectives.
With a specific configuration, the generator produces a consistent instance, ensuring its reproducibility.

\paragraph{Stage 2: optimal decision and workload definition.} 
For each family, a tailored exact algorithm is employed to solve every generated instance to global optimality, and the number of its executed primitive operations is used to define four workload levels.
In this way, we ensure that instances (from different families) at the same workload level require a comparable amount of computational effort.
\Cref{sec:workload-foundation} formalizes the primitive operations and workload levels.

\paragraph{Stage 3: tool-mediated world construction.}
For each family, we predefine a \emph{reusable world-construction template} that specifies the textual question $q$, the mappings from instance fields to tools $\mathcal{T}$, the tool-call costs $f_c$ and access budget $b$.
Instantiating the template with a hidden instance $\mathcal{I}$ produces two algorithmic worlds, one using the Direct interface $\mathcal{T}_{\mathrm{d}}$ and the other using the Mediated interface $\mathcal{T}_{\mathrm{m}}$.
The paired worlds share the same $\mathcal{I}$, $q$, $f_c$ and $b$; their difference lies in the relational structure used to present the instance information.
For each of the ten families and four workload levels, we generate three hidden instances and present each through both interfaces, yielding \(10\times4\times3=120\) hidden instances and \(120\times2=240\) algorithmic worlds.

\subsection{Data Quality}
\label{sec:data-quality}
To ensure data quality, we verify that each hidden instance possesses a feasible decision and a unique global optimum. Additionally, we confirm that sufficient information can be obtained using tools within the access budget.

\paragraph{Verified optima.} For each family, the construction algorithm is exact under the formal problem definition, but the optimal decision and objective value for a generated instance $\mathcal{I}$ still depend on the correctness of its implementation.
We therefore solve $\mathcal{I}$ a second time with an independently implemented exact solver based on a different state direction or problem presentation.
An instance is retained only if both implementations concur on its optimal decision, objective value, and identify the same unique optimum.

\paragraph{Information access.} For every algorithmic world $w$, we validate two alternative ways of gathering information ($\mathcal{T}_{\mathrm{d}}$ and $\mathcal{T}_{\mathrm{m}}$). Each way specifies exactly which information tools are called.
When these tools are invoked, the covered information must reconstruct $\mathcal{I}$ and recover its optimum, while their total cost remains within $b$.
We also evaluate the cost of exhaustively calling all tools to disclose instance information. The cost exceeds $b$ in all $w$, confirming that enough information can be acquired within the budget but exhaustive tool use cannot.

For more details of the validation, please refer to~\cref{app:certificates}.

\subsection{Evaluation Metrics}
\label{sec:metrics}
\paragraph{Final-decision metrics.} We use three metrics to evaluate the quality of the final decision $y$. If no final decision is submitted via $t_{\mathrm{dec}}$, all three metrics are set to zero.

(1) \emph{Feasibility.} For a world $w$, let $F_w(y)\in\{0,1\}$ indicate whether the decision $y$ is feasible. Specifically, $F_w(y)=1$ if $y$ satisfies all task constraints, and $F_w(y)=0$ otherwise.

(2) \emph{Exact optimality.} For a feasible decision $y$, let $v_w(y)$ denote its objective value and $v_w^\star$ the verified optimal value for $w$.
We define its optimality gap as $\Delta_w(y)=|v_w(y)-v_w^\star|$.
Exact optimality, our primary metric, is then defined as
\begin{equation}
E_w(y)=
\begin{cases}
1, & F_w(y)=1 \text{ and } \Delta_w(y)=0,\\
0, & \text{otherwise}.
\end{cases}
\label{eq:exact-optimality}
\end{equation}

(3) \emph{Reference Utility.} To assess the nuanced degree of optimality, we also define a reference optimality gap as $\Delta^{\mathrm{ref}}_w$, which is calculated based on a well-defined suboptimal reference decision.
In this way, the reference utility of $y$ is defined as:
\begin{equation}
U_w(y)=
\begin{cases}
\max\!\left\{0,1-\Delta_w(y) / \Delta^{\mathrm{ref}}_w \right\}, & F_w(y)=1,\\
0 ,& F_w(y)=0.
\end{cases}
\label{eq:reference-utility}
\end{equation}
$U_w(y)$ equals $1$ at the optimum, decreases to zero at the reference decision, and remains zero for feasible decisions no better than the reference or infeasible decisions.

\paragraph{Trajectory diagnostics.} We employ two diagnostic measures to evaluate the decision information gathered from $\mathcal{T}_{\mathrm{info}}$ along a trajectory $\tau$.
As described in \cref{sec:data-quality}, for each world $w$, we define the necessary information required to reconstruct $\mathcal{I}$.
Since this information can be acquired in various ways, a world may possess multiple sufficient information sets.
(1) \emph{Information sufficiency} records whether the gathered information in $\tau$ include a complete sufficient set, while (2) \emph{discovery coverage} measures the largest proportion of any sufficient set present within the gathered information.
Formally, let $\mathcal A_w(\tau)$ denote the set of instance information contained in $\tau$, and let $\mathcal S(w)$ denote the sufficient information sets described above. Information sufficiency and discovery coverage can be defined as:
\begin{equation}
C_w(\tau)=\mathds{1}\!\left[D_w(\tau)=1\right], \qquad D_w(\tau)=\max_{S\in\mathcal S(w)} \frac{|\mathcal A_w(\tau)\cap S|}{|S|}.
\label{eq:discovery-metrics}
\end{equation}
In this way, $C_w(\tau)=1$ and $E_w(y)=0$ indicate that the agent gathers enough information, but fail into a non-optimal decision.
Because trajectory diagnostics are separate from final-decision quality, they can be utilized to assess whether and how LLM agents turn available information into globally optimal decisions.
Further details on the sufficient information sets and diagnostic computation are provided in \cref{app:metrics}.
\section{Algorithmic Foundations}
\label{sec:foundations}

This section explains two algorithmic motivations that underlie the benchmark construction.
First, within each task family, individual decisions interact through shared constraints or objective terms, and a family-specific exact algorithm solves the resulting combinatorial optimization problem to global optimality.
Second, the primitive operations executed by these algorithms on each generated instance provide a reproducible construction scale for assigning instances to four workload levels.

\subsection{Global Coupling and Exact Algorithms}
\label{sec:foundations-exact}

In every family, actions are coupled through shared constraints or objectives.
Routing and sequencing families propagate residues or transition costs across successive actions, whereas other families couple actions through shared capacities, activation costs, or exact-cover conditions.
As a result, optimizing each action independently does not lead to a global optimum.
The family-specific exact algorithms therefore keep, at every step, the information that later actions depend on.
For example, in Fleet Dispatch, the dynamic program retains the vehicle-load vector while processing jobs: this vector records both how much room each vehicle still has and which activation costs have already been incurred, neither of which can be recovered from the total remaining capacity alone.

We run these exact algorithms only when constructing the benchmark, so that every instance carries a certified optimum and agent performance can be reported as a gap to it.
These procedures are never shown to the evaluated LLM agents, which must uncover the coupling structure of each family on their own and turn information acquired through tools into globally optimal decisions.

\subsection{Algorithm-Grounded Workload Levels}
\label{sec:workload-foundation}

Size configuration in deterministic instance generation does not provide a meaningful scale across families because their exact algorithms operate over different decision spaces.
For each generated instance, the number of five primitive operations (\emph{i.e.}, visited states, examined transitions, constraint checks, value updates, and reconstructed decision components) is counted:
\begin{equation}
  \operatorname{Work}(\mathcal{I})=
  N_{\mathrm{state}}+N_{\mathrm{transition}}+N_{\mathrm{check}}
  +N_{\mathrm{update}}+N_{\mathrm{reconstruct}}.
  \label{eq:algorithm-work}
\end{equation}
Every instance uses the same primary operations and counting rules across different families.
We assign family-specific configuration settings to four workload levels:
Levels 1-4 use the inclusive bands $[2^{15},2^{17}]$, $[2^{19},2^{21}]$, $[2^{22},2^{24}]$, and $[2^{24},2^{26}]$, respectively.
This procedure aligns instances $\mathcal{I}$ by the number of executed primitive operations of their exact algorithms without requiring their raw size configuration or decision-space sizes.
The gaps between consecutive bands are deliberate: they keep adjacent levels at least a factor of two apart in $\operatorname{Work}(\mathcal{I})$, so the levels stay clearly separable.
Levels 3 and 4 share the endpoint $2^{24}$ instead, and are separated by requiring $\operatorname{Work}(\mathcal{I})$ to increase strictly across levels within every family.
For more details of the operation-counting rules and the workloads implemented at each level, please refer to~\cref{app:work}.
\section{Experiments and Analysis}
\label{sec:experiments}

\subsection{Experimental Setup}
\label{sec:setup}

We evaluate seven leading LLMs: GPT-5.6 Terra and GPT-5.6 Sol \citep{openai2026gpt56}, Qwen 3.5 Plus \citep{qwen2026qwen35}, DeepSeek V4 Pro \citep{deepseek2026v4}, Claude Opus 4.8 and Claude Sonnet 5 \citep{anthropic2026opus48,anthropic2026sonnet5}, and GLM 5.2 \citep{zai2026glm52}.
We enable the reasoning mode for all LLM agents and limit the number of information-tool calls to a maximum of 255.
For each agent, we perform three trials and report the mean and sample standard deviation of the results.
Detailed hyperparameters used during model inference, such as temperature and reasoning effort, along with additional evaluation details, are provided in \cref{app:operations}.

\subsection{Main Results}
\label{sec:main-results}
\begin{table}[t]
\centering
\begingroup
\awtablestandard
\awtablezebra{3}
\resizebox{1.0\textwidth}{!}{%
\begin{tabular}{@{}lccccc@{}}
\toprule
\awtableheader
& \multicolumn{3}{c}{\textbf{Final-decision metrics}} & \multicolumn{2}{c}{\textbf{Trajectory diagnostics}} \\
\cmidrule(lr){2-4}\cmidrule(l){5-6}
\rowcolor{white}
Model & \shortstack{Exact\\optimality} & Feasibility & \shortstack{Reference\\utility} & \shortstack{Information\\sufficiency} & \shortstack{Discovery\\coverage} \\
\midrule
Qwen 3.5 Plus & 5.42 $\pm$ 0.42 & 48.89 $\pm$ 1.73 & 15.82 $\pm$ 0.87 & 37.92 $\pm$ 1.25 & 81.55 $\pm$ 0.42 \\
DeepSeek V4 Pro & 16.11 $\pm$ 1.05 & 83.89 $\pm$ 1.34 & 28.49 $\pm$ 0.91 & 79.58 $\pm$ 1.67 & 97.74 $\pm$ 0.11 \\
GLM 5.2 & 24.86 $\pm$ 0.64 & 64.58 $\pm$ 3.15 & 35.48 $\pm$ 2.67 & 83.33 $\pm$ 3.82 & 93.91 $\pm$ 1.49 \\
GPT-5.6 Terra & 26.39 $\pm$ 1.34 & 87.78 $\pm$ 0.87 & 54.10 $\pm$ 1.95 & 79.72 $\pm$ 3.13 & 96.65 $\pm$ 0.25 \\
GPT-5.6 Sol & 38.19 $\pm$ 1.34 & \textbf{97.50 $\pm$ 0.00} & \textbf{74.98 $\pm$ 1.77} & \textbf{89.86 $\pm$ 0.64} & \textbf{99.10 $\pm$ 0.13} \\
Claude Sonnet 5 & 32.36 $\pm$ 3.85 & 92.08 $\pm$ 0.83 & 67.20 $\pm$ 9.53 & 86.39 $\pm$ 7.36 & 98.59 $\pm$ 1.08 \\
Claude Opus 4.8 & \textbf{38.61 $\pm$ 4.65} & 96.25 $\pm$ 2.89 & 73.14 $\pm$ 11.81 & 81.53 $\pm$ 8.74 & 97.33 $\pm$ 1.45 \\
\bottomrule
\end{tabular}%
}
\caption{Performance on all 240 algorithmic worlds in \algoworlds. Entries are
the mean $\pm$ sample standard deviation over three trials. \textbf{Bold} marks the best performance.}
\label{tab:main-results}
\endgroup
\end{table}

\paragraph{Optimality vs. Feasibility.} \Cref{tab:main-results} presents the overall performance of all LLM agents. Notably, none of the agents achieve an exact-optimality score exceeding 40\%.
The best LLM, Claude Opus 4.8, achieves 38.61\%, followed closely by GPT-5.6 Sol at 38.19\%.
In terms of feasibility, the scores are substantially higher: Claude Opus 4.8 and GPT-5.6 Sol achieve 96.25\% and 97.50\%, respectively.
Consequently, in over half of the cases, these two leading LLMs produce decisions that are feasible but not optimal, demonstrating that the most advanced models are significantly more adept at meeting task constraints than at identifying the global optimum.
This result also highlights the challenges within our \algoworlds, suggesting that making a globally optimal decision remains an open question for LLM agents.

\begin{table}[t]
\centering
\begingroup
\awtablestandard
\awtablezebra{3}
\resizebox{1.0\textwidth}{!}{%
\begin{tabular}{lcccc}
\toprule
\awtableheader
& \makecell{\textbf{Information}\\\textbf{sufficiency}} & \multicolumn{3}{c}{\textbf{Three final results in informative trajectories}} \\
\cmidrule(lr){3-5}
\rowcolor{white}
Model & & Globally optimal & Feasible but suboptimal & Other failure \\
\midrule
Qwen 3.5 Plus & 37.9 & 3.3 & 49.9 & 46.8 \\
DeepSeek V4 Pro & 79.6 & 17.8 & 69.4 & 12.8 \\
GLM 5.2 & 83.3 & 26.0 & 40.5 & 33.5 \\
GPT-5.6 Terra & 79.7 & 29.2 & 59.8 & 11.0 \\
GPT-5.6 Sol & 89.9 & 41.6 & 56.3 & 2.1 \\
Claude Sonnet 5 & 86.4 & 33.9 & 58.0 & 8.1 \\
Claude Opus 4.8 & 81.5 & 36.3 & 59.5 & 4.2 \\
\bottomrule
\end{tabular}%
}
\caption{Information sufficiency and the final results in informative trajectories.}
\label{tab:conditional-outcomes}
\endgroup
\end{table}

\paragraph{Informative but Suboptimal.} The trajectory diagnostics further show that the gap between optimality and feasibility \emph{cannot} be explained by information acquisition alone.
The information-sufficiency score exceeds the exact-optimality score for every model, ranging from 37.92\% to 89.86\%.
Discovery coverage is higher still: five of the seven models exceed 96\%, with GPT-5.6 Sol reaching 99.10\%.
The results indicate that agents generally gather enough information to determine the optimum but still fail to turn that information into a globally optimal decision.

To further investigate this gap, we categorize trajectories with sufficient information (denoted as \emph{informative trajectories}) into three distinct final results: globally optimal, feasible but suboptimal, and other failures (\emph{e.g.}, infeasible decisions or absent final decisions).
As shown in \Cref{tab:conditional-outcomes}, only 29.2\% of GPT-5.6 Terra's informative trajectories achieve a globally optimal outcome, while the remaining 70.8\% fall short, with 59.8\% resulting in feasible but suboptimal decisions.
This pattern is consistent across all seven models, with ``feasible but suboptimal'' representing the largest proportion among the three results.
Even for GPT-5.6 Sol, 56.3\% of informative trajectories result in feasible but suboptimal decisions, while only 41.6\% reach the optimum.
Information sufficiency measures the gathered information in trajectories, rather than the agent's ability to recognize or utilize it correctly.
This finding thus underscores the gap between the ability to gather information and the capacity to reason towards an optimal decision in \algoworlds.
Detailed conditional outcome counts and structured trajectory examples appear in \cref{tab:conditional-outcome-counts,app:trajectories}.

\subsection{Performance across Different Perspectives}
\label{sec:task-analysis}

We analyze the model performance specifically from the three perspectives of task families, workload levels, and tool interfaces:

\paragraph{Task families.}
\begin{table}[h]
\centering
\begingroup
\awtablestandard
\setlength{\tabcolsep}{9pt}
\awtablezebra{2}
\resizebox{1.0\textwidth}{!}{%
\begin{tabular}{lccccccc}
\toprule
\awtableheader
Task family & Terra & Sol & Qwen & DeepSeek & Opus & Sonnet & GLM \\
\midrule
Transit Routing & \cellcolor{AWteal!4}0.0 & \cellcolor{AWteal!4}\textbf{1.4} & \cellcolor{AWteal!4}0.0 & \cellcolor{AWteal!4}0.0 & \cellcolor{AWteal!4}0.0 & \cellcolor{AWteal!4}0.0 & \cellcolor{AWteal!4}0.0 \\
Basket Assembly & \cellcolor{AWteal!28}\textbf{100.0} & \cellcolor{AWteal!28}\textbf{100.0} & \cellcolor{AWteal!15}47.2 & \cellcolor{AWteal!25}88.9 & \cellcolor{AWteal!28}\textbf{100.0} & \cellcolor{AWteal!28}\textbf{100.0} & \cellcolor{AWteal!27}97.2 \\
Station Siting & \cellcolor{AWteal!24}83.3 & \cellcolor{AWteal!28}\textbf{98.6} & \cellcolor{AWteal!4}0.0 & \cellcolor{AWteal!6}8.3 & \cellcolor{AWteal!27}97.2 & \cellcolor{AWteal!26}90.3 & \cellcolor{AWteal!18}58.3 \\
Authorization Planning & \cellcolor{AWteal!8}18.1 & \cellcolor{AWteal!21}\textbf{72.2} & \cellcolor{AWteal!4}0.0 & \cellcolor{AWteal!5}4.2 & \cellcolor{AWteal!19}62.5 & \cellcolor{AWteal!18}59.7 & \cellcolor{AWteal!9}19.4 \\
Series Portfolio & \cellcolor{AWteal!4}0.0 & \cellcolor{AWteal!4}0.0 & \cellcolor{AWteal!4}0.0 & \cellcolor{AWteal!4}0.0 & \cellcolor{AWteal!6}\textbf{9.7} & \cellcolor{AWteal!4}1.4 & \cellcolor{AWteal!4}0.0 \\
Machine Layout & \cellcolor{AWteal!4}1.4 & \cellcolor{AWteal!4}1.4 & \cellcolor{AWteal!4}0.0 & \cellcolor{AWteal!4}0.0 & \cellcolor{AWteal!5}\textbf{4.2} & \cellcolor{AWteal!4}0.0 & \cellcolor{AWteal!4}0.0 \\
Sequential Matching & \cellcolor{AWteal!4}1.4 & \cellcolor{AWteal!6}\textbf{6.9} & \cellcolor{AWteal!4}0.0 & \cellcolor{AWteal!4}0.0 & \cellcolor{AWteal!4}0.0 & \cellcolor{AWteal!5}2.8 & \cellcolor{AWteal!4}1.4 \\
Fleet Dispatch & \cellcolor{AWteal!18}59.7 & \cellcolor{AWteal!27}\textbf{94.4} & \cellcolor{AWteal!6}6.9 & \cellcolor{AWteal!18}59.7 & \cellcolor{AWteal!25}87.5 & \cellcolor{AWteal!16}50.0 & \cellcolor{AWteal!21}72.2 \\
Evidence-Joined Routing & \cellcolor{AWteal!4}0.0 & \cellcolor{AWteal!4}0.0 & \cellcolor{AWteal!4}0.0 & \cellcolor{AWteal!4}0.0 & \cellcolor{AWteal!5}2.8 & \cellcolor{AWteal!5}\textbf{4.2} & \cellcolor{AWteal!4}0.0 \\
Migration Portfolio & \cellcolor{AWteal!4}0.0 & \cellcolor{AWteal!6}6.9 & \cellcolor{AWteal!4}0.0 & \cellcolor{AWteal!4}0.0 & \cellcolor{AWteal!9}\textbf{22.2} & \cellcolor{AWteal!8}15.3 & \cellcolor{AWteal!4}0.0 \\
\bottomrule
\end{tabular}%
}
\caption{Performance across task families (in terms of exact optimality). The pale teal color represents the magnitude of the scores, and the \textbf{bold} indicates the row-best value.}
\label{tab:task-results}
\endgroup
\end{table}

\Cref{tab:task-results} illustrates the significant variation in exact-optimality scores across different task families. For instance, the scores for Station Siting, Fleet Dispatch, and Authorization Planning range from 0.0\% to 98.6\%, 6.9\% to 94.4\%, and 0.0\% to 72.2\% across various LLMs, respectively. Notably, Basket Assembly achieves the highest scores, with four LLMs reaching 100\% optimality. In stark contrast, no model surpasses 10\% in tasks such as Transit Routing, Series Portfolio, Machine Layout, Sequential Matching, and Evidence-Joined Routing.

The disparity between Transit Routing and Basket Assembly is particularly revealing. Although the exact algorithm for Transit Routing has a worst-case bound that is polynomial in its listed parameters, the bound for Basket Assembly includes an exponential factor, as detailed in \cref{app:cases}.
Despite this, LLMs perform significantly better on Basket Assembly. This suggests that the asymptotic complexity of the benchmark's exact algorithms alone does not account for the observed differences in performance across task families.

\paragraph{Workload levels.}

\begin{table}[t]
\centering
\begingroup
\awtablestandard
\setlength{\tabcolsep}{10.5pt}
\awtablezebra{3}
\resizebox{1.0\textwidth}{!}{
\begin{tabular}{lcccc}
\toprule
\awtableheader
& \multicolumn{4}{c}{\textbf{Workload level}} \\
\cmidrule(lr){2-5}
\rowcolor{white}
Model & L1 & L2 & L3 & L4 \\
\midrule
GPT-5.6 Terra & 27.2 / 68.4 & 30.6 / 61.0 & 21.7 / 47.2 & 26.1 / 39.8 \\
GPT-5.6 Sol & 42.2 / 87.8 & 38.9 / 81.3 & 35.0 / 69.2 & 36.7 / 61.7 \\
Qwen 3.5 Plus & 4.4 / 20.3 & 10.0 / 17.1 & 4.4 / 15.3 & 2.8 / 10.6 \\
DeepSeek V4 Pro & 19.4 / 35.9 & 18.3 / 27.6 & 11.7 / 22.3 & 15.0 / 28.1 \\
Claude Opus 4.8 & 44.4 / 84.0 & 41.1 / 76.7 & 32.2 / 69.8 & 36.7 / 62.1 \\
Claude Sonnet 5 & 36.1 / 77.1 & 36.1 / 73.9 & 27.2 / 61.3 & 30.0 / 56.5 \\
GLM 5.2 & 26.7 / 40.5 & 26.7 / 37.2 & 20.6 / 32.2 & 25.6 / 32.1 \\
\bottomrule
\end{tabular}
}
\caption{Performance across workload levels (in terms of exact optimality / reference utility).}
\label{tab:workload-results}
\endgroup
\end{table}

\Cref{tab:workload-results} shows an overall decline in decision performance as the workload level increases.
Reference utility decreases at every successive level for six of the seven models and is lower at L4 than at L1 for all seven, with overall declines of about 8\textasciitilde29\%.
Exact optimality follows the same general pattern: every model achieves a lower rate at L4 than at L1, with declines of 1.1\textasciitilde7.7\%.
Because the workload levels are assigned according to the primitive operations executed by the exact algorithms, independently of model trajectories, this shared trend indicates that the algorithm-grounded scale captures a meaningful dimension of challenge for current LLM agents.

\paragraph{Tool Interfaces.}

\begin{table}[t]
\centering
\begingroup
\awtablestandard
\setlength{\tabcolsep}{6pt}
\awtablezebra{3}
\resizebox{0.95\textwidth}{!}{%
\begin{tabular}{@{}lccccccc@{}}
\toprule
\awtableheader
& \multicolumn{4}{c}{\textbf{Exact-optimality comparison}} & \multicolumn{3}{c}{\textbf{Matched outcomes}} \\
\cmidrule(lr){2-5}\cmidrule(lr){6-8}
\rowcolor{white}
\textbf{Model} & \textbf{Direct} & \textbf{Mediated} & \textbf{$\Delta_{\mathrm{D-M}}$} & \makecell{\textbf{95\% paired}\\\textbf{bootstrap interval}} & \makecell{\textbf{Direct}\\\textbf{wins}} & \makecell{\textbf{Mediated}\\\textbf{wins}} & \textbf{Ties} \\
\midrule
GPT-5.6 Terra & 27.2 & 25.6 & +1.7 & $[-1.9, +5.3]$ & 19 & 13 & 328 \\
GPT-5.6 Sol & 40.6 & 35.8 & +4.7 & $[+1.4, +8.3]$ & 25 & 8 & 327 \\
Qwen 3.5 Plus & 6.1 & 4.7 & +1.4 & $[-0.8, +3.9]$ & 11 & 6 & 343 \\
DeepSeek V4 Pro & 16.4 & 15.8 & +0.6 & $[-1.9, +3.1]$ & 15 & 13 & 332 \\
Claude Opus 4.8 & 39.2 & 38.1 & +1.1 & $[-1.7, +3.9]$ & 17 & 13 & 330 \\
Claude Sonnet 5 & 35.6 & 29.2 & +6.4 & $[+2.2, +11.1]$ & 34 & 11 & 315 \\
GLM 5.2 & 26.1 & 23.6 & +2.5 & $[-0.8, +6.1]$ & 20 & 11 & 329 \\
\bottomrule
\end{tabular}%
}
\caption{Matched Direct--Mediated exact-optimality results. Direct and Mediated
are mean rates (\%) across three trials, and $\Delta_{\mathrm{D-M}}$ is their
difference in percentage points. Direct wins, Mediated wins, and ties count the
360 matched instance--trial pairs for each model. The intervals are 95\% paired
bootstrap confidence intervals for $\Delta_{\mathrm{D-M}}$; complete statistical
details appear in \cref{app:paired-interface-results}.}
\label{tab:representation-results}
\endgroup
\end{table}

\Cref{tab:representation-results} compares exact-optimality scores under the Direct and Mediated interfaces.
The Direct-minus-Mediated difference, denoted as $\Delta_{\mathrm{D-M}}$, is positive for all seven models, ranging from +0.6\% to +6.4\%.
Despite these aggregate differences, 315--343 of the 360 paired evaluations for each model have the same exact-optimality status under both interfaces.
The 95\% confidence interval for $\Delta_{\mathrm{D-M}}$ excludes zero only for GPT-5.6 Sol and Claude Sonnet 5.
Overall, exact-optimality outcomes are broadly stable across the paired interfaces, with a distinguishable Direct advantage for these two models. Additional experimental results are provided in \cref{app:full-results}.
\section{Limitations}
\label{sec:limitation}

While we show the challenges lie in \algoworlds, there are some limitations worth noting:
(1) Our experiments are based on seven leading LLMs, and the results may reflect inherent biases present in these models.
Besides, the results only characterize LLMs with information tools, and providing them with more external tools (\emph{e.g.}, a general-purpose code execution environment or an external optimization solver) might enhance their performance in \algoworlds.
(2) All worlds are formally generated and have unique global optima. This approach allows for precise verification; however, it fails to account for the noise, ambiguity, and temporal fluctuations inherent in real-world information sources, as well as decision problems that may have multiple globally optimal solutions.
(3) The trajectory diagnostics assess the gathered information against sufficient information sets. However, these sets may not encompass all possible methods for determining the optimum. As a result, a trajectory that does not fully include any of these sets might still possess enough information to facilitate a globally optimal decision.

\section{Conclusion}
\label{sec:conclusion}

We introduce \algoworlds, a benchmark for evaluating whether LLM agents can turn information acquired through tools into globally optimal decisions.
It comprises 120 hidden optimization instances spanning ten formally specified combinatorial optimization families and four workload levels.
Each instance has an independently verified unique global optimum and is presented through paired Direct and Mediated interfaces, yielding 240 algorithmic worlds.
Our extensive experiments with seven leading LLMs reveal that the top-performing model, Claude Opus 4.8, achieves an exact-optimality score of only \ResultBestExact, despite demonstrating significantly higher feasibility. This suggests that even when LLMs gather sufficient information to identify the global optimum, most still fail to achieve it.
These findings highlight that acquiring adequate information and adhering to task constraints do not necessarily guarantee consistent global optimization. The challenge of making globally optimal decisions remains unresolved for LLM agents.

\bibliography{reference}

\newpage
\appendix
\onecolumn
\crefalias{section}{appendix}
\crefalias{subsection}{appendix}
\section{A Worked Fleet Dispatch Pair}
\label{app:worked-world}

This section gives a concrete view of one released Fleet Dispatch pair at
workload level 1 and instance index 1. It focuses on what the agent receives,
how the paired tools present the hidden information, and how the agent submits
its final decision. We
denote its Direct and Mediated worlds by $w^{(\mathrm D)}$ and
$w^{(\mathrm M)}$. The shared instance contains 14 jobs and five vehicles. For
readability, we replace its opaque public identifiers with
$J_1,\ldots,J_{14}$ and $V_1,\ldots,V_5$; all numerical values and relations
are unchanged. Repeated records are omitted.

During evaluation, the agent receives the textual question, tool schemas and
costs, and access budget, but not the hidden instance or validation results.
The example distinguishes this model-visible content from the offline records
used for construction and validation.

\paragraph{World components and visibility.}
The shared hidden instance $\mathcal I$ contains the job loads, eligible
vehicle--job pairs, vehicle capacities, route costs, and fixed activation
costs. This instance has 14 positive job loads, five vehicles with capacity 4,
five activation costs, and a $5\times14$ route-cost matrix. The eligibility
relation determines which matrix entries correspond to admissible assignments.
For example, $J_1$ has load 1, the fixed activation cost of $V_1$ is
122,070,312,500, and the route cost from $V_1$ to $J_1$ is 6,103,515,625.
These values belong to the hidden instance and become visible only when the
corresponding information-tool calls return them. Interface-specific
identifiers and serialized response records are not part of $\mathcal I$.

The textual question $q$ lists the public job and vehicle identifiers. It asks
the agent to assign every job to an eligible vehicle without exceeding vehicle
capacity and to minimize the Fleet Dispatch objective $g_{\mathcal I}$ defined
in \cref{app:cases}. This objective is the sum of route costs,
activation costs, and the decision-independent offset
$\gamma_{\mathcal I}$. The offset changes the reported objective value but
not the minimizing mapping. The agent also receives the tool schemas in
$\mathcal T$, their per-call costs under $f_c$, and the access budget $b=97$.
The final-decision tool requires a 14-entry job-to-vehicle mapping.

\Cref{tab:worked-world-tools} summarizes the available tools and their costs
in the paired worlds.

\begin{table}[H]
\centering
\begingroup
\awtablecompact
\awtablezebra{2}
\resizebox{0.90\textwidth}{!}{%
\begin{tabular}{@{}
  >{\raggedright\arraybackslash}p{4.0cm}
  >{\raggedright\arraybackslash}p{2.75cm}
  >{\raggedright\arraybackslash}p{6.5cm}
  >{\centering\arraybackslash}p{1.0cm}@{}}
\toprule
\awtableheader
Tool & Arguments & Returned information or effect & $f_c(t)$ \\
\midrule
\texttt{read\_fleet\_policy} & None & Vehicle capacities and the complete
vehicle and job rosters & 2 \\
\texttt{read\_job} & One job & Its load and eligible vehicles, without route
costs & 1 \\
\texttt{read\_vehicle\_contract} & One vehicle & Its fixed activation cost,
without capacity or assigned jobs & 4 \\
\texttt{read\_route\_row} & One vehicle & Interface-specific route records
associated with that vehicle & 3 \\
\texttt{read\_dispatch\_window} & Start job and length 2 or 3 & Loads,
eligible vehicles, and interface-specific route records for consecutive jobs
& 14 \\
\texttt{evaluate\_assignment} & One complete mapping & Feasibility and, when
feasible, its objective value & 5 \\
\texttt{submit\_assignment} & One complete mapping & Records the binding final
decision and ends the trajectory & 0 \\
\bottomrule
\end{tabular}%
}
\caption{The available tools in the worked Fleet Dispatch pair. The first
five reveal instance information. \texttt{evaluate\_assignment} provides
nonterminal feedback and may be used once, whereas
\texttt{submit\_assignment} records the binding final decision.}
\label{tab:worked-world-tools}
\endgroup
\end{table}

\paragraph{Representative Direct and Mediated responses.}
Calling \texttt{read\_job} on $J_1$ returns its load and the five eligible
vehicles. In the Direct world, a route-row call on $V_1$ returns a record that
includes the corresponding route cost:
\begin{verbatim}
read_job({job: J1})
  -> {job_id: J1, load: 1, eligible_vehicles: [V1, ..., V5]}

Direct route record:
  {vehicle_id: V1, job_id: J1, route_cost: 6103515625}
\end{verbatim}
The Mediated world has the same textual question, tool names, argument schemas,
costs, and budget. Its route handler exposes the same route fact through an
intermediate shift:
\begin{verbatim}
Mediated route records:
  {vehicle_id: V1, shift_id: S1}
  {shift_id: S1, job_id: J1, route_cost: 6103515625}
\end{verbatim}
Joining the two Mediated records on $S_1$ and then removing that
interface-specific identifier yields the same route-cost fact as the Direct
record. Both presentations therefore encode
$(V_1,J_1,6{,}103{,}515{,}625)$. The identifier $S_1$ appears in the Mediated
response but is not part of $\mathcal I$. The general response-normalization
procedure and the pair-level equivalence check are defined in
\cref{app:acquisition,app:paired-equivalence}.

\paragraph{Candidate evaluation and the binding decision.}
The \texttt{evaluate\_assignment} tool accepts one complete mapping, costs five
units, and may be called at most once. It reports feasibility and, for a
feasible mapping, its objective value. The call is nonterminal and does
not commit a final decision. Because it evaluates an agent-proposed mapping
rather than revealing fields of the hidden instance, its feedback is not
counted as instance information by the trajectory diagnostics.

Only \texttt{submit\_assignment} is the final-decision tool $t_{\mathrm{dec}}$.
It accepts one 14-entry object mapping every $J_i$ to a $V_j$, costs zero, and
terminates the trajectory. The family checker verifies that every job appears
exactly once, every selected vehicle--job pair is eligible, and no vehicle
capacity is exceeded. For a feasible mapping, it recomputes the
objective $g_{\mathcal I}$. The submitted mapping is the final decision $y$
scored by the metrics in \cref{sec:metrics}. A mapping previously passed to the
candidate-evaluation tool is not binding unless it is later submitted through
$t_{\mathrm{dec}}$.

\section{Formal Problems and Exact Algorithms}
\label{app:cases}

This section formalizes the ten optimization families and describes the
exact algorithms used during benchmark construction. For each family, we give
the decision, feasibility conditions, objective, source of global coupling,
algorithm, correctness argument, and complexity. The independent verification
algorithms and the remaining validation procedures are described in
\cref{app:certificates}.

\subsection{Transit Routing}

\paragraph{Problem and global coupling.}
The instance has a positive residue modulus $M$ and a target residue
$r_\star\in\{0,\ldots,M-1\}$. It contains $d$ ordered stages, each with a set
$E_i$ of candidate legs. Let
$\mathcal L=\{\ell(e):e\in\bigcup_i E_i\}$ be the set of lines. A leg $e$ has
travel time $\tau(e)$, line $\ell(e)\in\mathcal L$, and residue
$\rho(e)\in\{0,\ldots,M-1\}$. A route
$x=(e_1,\ldots,e_d)$ selects one leg from each stage and is feasible when
\begin{equation}
  \sum_{i=1}^{d}\rho(e_i)\equiv r_\star\pmod M,
  \qquad e_i\in E_i.
\end{equation}
Its objective is
\begin{equation}
  \min_x\
  \sum_{i=1}^{d}\tau(e_i)
  +\sum_{i=2}^{d}
    C\!\left(\ell(e_{i-1}),\ell(e_i)\right),
\end{equation}
where $C(\lambda,\lambda')$ is the directed transfer cost for an ordered pair
of lines and need not be symmetric. The
modular condition couples the residues of all selected legs, while each
transfer cost couples two consecutive stages. A route prefix must therefore
retain both its accumulated residue and its final line.

\paragraph{Exact algorithm and correctness.}
Let $D_i(\lambda,r)$ be the minimum cost of a prefix through stage $i$ whose
final line is $\lambda$ and whose accumulated residue is $r$ modulo $M$.
Initialize every first-layer value to $+\infty$ and, for each $e\in E_1$, set
\begin{equation}
  D_1\!\left(\ell(e),\rho(e)\right)
  \leftarrow
  \min\!\left\{D_1\!\left(\ell(e),\rho(e)\right),\tau(e)\right\}.
\end{equation}
For $i\geq2$, each reachable state $(\lambda,r)$ at stage $i-1$ and each leg
$e\in E_i$ induce the update
\begin{equation}
\begin{aligned}
  D_i\!\left(\ell(e),(r+\rho(e))\bmod M\right)
  \leftarrow\min\Bigl\{&
  D_i\!\left(\ell(e),(r+\rho(e))\bmod M\right),\\
  &D_{i-1}(\lambda,r)+\tau(e)+C\!\left(\lambda,\ell(e)\right)
  \Bigr\}.
\end{aligned}
\end{equation}
The optimum is $\min_{\lambda\in\mathcal L}D_d(\lambda,r_\star)$; stored
predecessor states and legs reconstruct the route.

The state is sufficient because an earlier prefix can affect a continuation
only through its final line and accumulated residue. Induction over the stages
therefore shows that retaining the cheapest prefix for each state preserves an
optimal complete route.

The construction algorithm evaluates the same recurrence using ordered
predecessor scans. The ordering changes how candidate transitions are examined
but not the recurrence. In the benchmark construction,
$|\mathcal L|=p=k+2$ and $|E_i|=k$ for every stage. Including the
$O(p^2\log p)$ ordering step, the algorithm runs in
$O(p^2\log p+dk^2M)$ time.

\subsection{Basket Assembly}

\paragraph{Problem and global coupling.}
Let $A$, $B$, and $C$ be pairwise disjoint item groups with
$|A|=|B|=|C|=n$, and let $U=A\cup B\cup C$. A finite catalog $\mathcal E$
contains tickets $e$ with prices $p_e$; each ticket is either a singleton or a
triple containing one item from each group. A decision $X\subseteq\mathcal E$
is feasible when it forms an exact cover:
\begin{equation}
  \sum_{e\in X}\mathds{1}[v\in e]=1
  \qquad\text{for every }v\in U.
\end{equation}
The objective is to minimize $\sum_{e\in X}p_e$. Tickets interact through the
exact-cover constraints: selecting one ticket simultaneously covers all of its
items and excludes every other ticket containing any of them.

\paragraph{Exact algorithm and correctness.}
Fix an order on $U$, and represent each subset by a bitmask. For
$S\subseteq U$ representing the items already covered exactly once, let
$F(S)$ be the minimum additional price of tickets that are disjoint from $S$
and cover every item in $U\setminus S$ exactly once. Set $F(U)=0$. For
$S\subsetneq U$, let $u(S)$ be the least-indexed uncovered item. The recurrence
is
\begin{equation}
  F(S)=
  \min_{\substack{e\in\mathcal E,\ u(S)\in e\\e\cap S=\varnothing}}
  \left\{p_e+F(S\cup e)\right\},
  \label{eq:basket-dp}
\end{equation}
where a state with no admissible ticket has value $+\infty$. Every exact-cover
completion of $S$ contains exactly one admissible ticket covering $u(S)$.
Conditioning on that ticket gives \cref{eq:basket-dp}; induction on
$|U\setminus S|$ therefore establishes that $F(S)$ is the minimum remaining
cost. Stored minimizing tickets reconstruct the selected ticket set.

Let $\Delta$ be the maximum number of candidate tickets examined at one
reached state. The algorithm runs in $O(2^{3n}\Delta)$ time.

\subsection{Station Siting}

\paragraph{Problem and global coupling.}
There are $n$ ordered zones and two site options, labeled $0$ and $1$, per
zone. Option $a$ at zone $i$ has cost $\kappa_i(a)$. A decision
$x\in\{0,1\}^n$ must satisfy the shared capacity constraint
$\sum_i x_i\leq H$. Let
$\mathcal D\subseteq\{(i,j):1\leq i<j\leq n,\ j-i\leq\omega\}$ be the set of
demand relations. Each $(i,j)\in\mathcal D$ contributes reward $r_{ij}$
exactly when its endpoint choices equal a required pair
$(\alpha_{ij},\beta_{ij})\in\{0,1\}^2$. The objective is
\begin{equation}
  \max_{x\in\{0,1\}^n}
  \left\{
    \sum_{(i,j)\in\mathcal D}r_{ij}
      \mathds{1}\!\left[(x_i,x_j)
      =(\alpha_{ij},\beta_{ij})\right]
    -\sum_{i=1}^{n}\kappa_i(x_i)
  \right\}
  \quad\text{s.t.}\quad
  \sum_{i=1}^{n}x_i\leq H.
\end{equation}
All option-one choices consume the same capacity, and each demand reward
depends jointly on two zone choices. These shared terms prevent the zones from
being optimized independently.

\paragraph{Exact algorithm and correctness.}
Every optimal decision in the generated instances uses exactly $H$ option-one
sites. This construction property is checked by the independent verification
algorithm, which retains the original at-most-$H$ capacity constraint
(\cref{tab:verification-methods}). Writing $\bar H=n-H$, the construction
algorithm can therefore search decisions with exactly $\bar H$ option-zero
choices without changing the optimal value. The algorithm processes zones
from left to right and records the number of option-zero
choices used together with the most recent $\omega$ binary choices. The empty
prefix has value zero and uses no option-zero choices. Appending a choice
subtracts its site cost, increments the counter when that choice is zero, and
closes every demand term whose right endpoint is the new zone. States that
cannot be extended to exactly $\bar H$ option-zero choices are discarded, and
the terminal state must have counter $\bar H$.

Choices older than the frontier cannot occur in a future demand term because
of the span bound. The counter retains the past information needed by the
shared capacity constraint under this construction property. Each site cost is
subtracted when its zone is processed, and each demand reward is added once
when its right endpoint enters the frontier. Induction over the zones shows
that the best value for each state preserves an optimal completion;
backpointers reconstruct the site choices.

The algorithm runs in
$O(n(\bar H+1)2^\omega(\omega+1))$ time and
$O((\bar H+1)2^\omega)$ value-only memory.

\subsection{Authorization Planning}

\paragraph{Problem and global coupling.}
The instance has a positive residue modulus $M$, a target residue
$R\in\{0,\ldots,M-1\}$, $d$ ordered stages, a package set $\mathcal V_i$ at
each stage, and authorized transitions
$\mathcal E_i\subseteq\mathcal V_i\times\mathcal V_{i+1}$ between consecutive
stages. A decision is a sequence $x=(v_1,\ldots,v_d)$ with
$v_i\in\mathcal V_i$. It is feasible when
$(v_i,v_{i+1})\in\mathcal E_i$ for every $i<d$ and
\begin{equation}
  \sum_{i=1}^{d}z(v_i)\equiv R\pmod M,
\end{equation}
where $z(v)\in\{0,\ldots,M-1\}$ is the clearance increment of package $v$.
Each package also has a base price $p(v)$ and a policy code
$\operatorname{code}(v)$, whose adjustment is
$\eta[\operatorname{code}(v)]$. Let $h(u,v)$ be the handoff cost of an
authorized transition. The objective is
\begin{equation}
  \min_x\
  \sum_{i=1}^{d}
    \left(p(v_i)+\eta[\operatorname{code}(v_i)]\right)
  +\sum_{i=1}^{d-1}h(v_i,v_{i+1}).
\end{equation}
The preceding package determines the feasibility and handoff cost of the next
selection, while the modular condition couples the clearance increments
across the complete sequence.

\paragraph{Exact algorithm and correctness.}
Let $D_i(v,r)$ be the minimum cost of a feasible prefix through stage $i$ that
ends at package $v\in\mathcal V_i$ and has accumulated clearance residue $r$
modulo $M$. Initialize all values to $+\infty$ and, for every
$v\in\mathcal V_1$, set
\begin{equation}
  D_1(v,z(v))=p(v)+\eta[\operatorname{code}(v)].
\end{equation}
For each $(u,v)\in\mathcal E_i$ and residue $r$, the transition is
\begin{equation}
\begin{aligned}
  D_{i+1}\!\left(v,(r+z(v))\bmod M\right)
  \leftarrow\min\Bigl\{&
  D_{i+1}\!\left(v,(r+z(v))\bmod M\right),\\
  &D_i(u,r)+p(v)+\eta[\operatorname{code}(v)]+h(u,v)
  \Bigr\}.
\end{aligned}
\end{equation}
The optimum is $\min_{v\in\mathcal V_d}D_d(v,R)$; stored predecessors
reconstruct the package sequence.

The state is sufficient because the final package determines the authorization
and handoff effects on the next stage. The accumulated residue is the only
other prefix information needed to enforce the terminal residue condition. The
transition enumerates every authorized extension and adds each package and
handoff term exactly once. Induction over the stages therefore establishes that
the minimum terminal value is the global optimum. Let
$V=\sum_{i=1}^{d}|\mathcal V_i|$ and
$E=\sum_{i=1}^{d-1}|\mathcal E_i|$. The algorithm runs in
$O(M(V+E))$ time.

\subsection{Series Portfolio}

\paragraph{Problem and global coupling.}
Each of $n$ titles has two lots, represented by a binary choice
$x_i\in\{0,1\}$, where lot $z$ of title $i$ has value $v_{i,z}$. Let
$\mathcal F$ be a finite set of signed factors. Each $s\in\mathcal F$ has a
nonempty scope $P_s\subseteq\{1,\ldots,n\}$ satisfying
$\max P_s-\min P_s+1\leq\omega$ and a complete table
$A_s:\{0,1\}^{|P_s|}\rightarrow\mathbb Z$. The objective is
\begin{equation}
  \max_{x\in\{0,1\}^n}
  \left\{
    \sum_{i=1}^{n}v_{i,x_i}
    -C_{\mathrm{title}}
    +\sum_{s\in\mathcal F} A_s(x|_{P_s})
  \right\},
  \qquad
  C_{\mathrm{title}}
  =\sum_{i=1}^{n}\min_{z\in\{0,1\}}v_{i,z}.
\end{equation}
The offset $C_{\mathrm{title}}$ is independent of $x$ and therefore does not
change the maximizing portfolio. Overlapping factor scopes make the value of
one title depend on choices at several nearby titles.

\paragraph{Exact algorithm and correctness.}
The construction algorithm uses suffix states. Let $B_i(\sigma)$ be the
maximum sum of the first $i$ lot values and all factor values whose scopes end
at or before title $i$, among prefixes whose latest
$\min\{i,\omega-1\}$ choices form suffix $\sigma$.
Initialize $B_0(\varnothing)=0$ and all other values to $-\infty$. To advance
from title $i-1$ to title $i$, the algorithm appends a choice
$z\in\{0,1\}$, adds $v_{i,z}$ and every factor $s\in\mathcal F$ with
$\max P_s=i$, and maximizes into the state retaining the latest
$\omega-1$ choices. Each factor is thus added exactly once, when its complete
scope is known. The optimum is
$\max_\sigma B_n(\sigma)-C_{\mathrm{title}}$.

Because every scope has span at most $\omega$, no future factor can involve a
choice older than the retained suffix. Induction over title positions
therefore establishes that the best value for each suffix state preserves a
globally optimal completion; stored predecessors reconstruct the binary
portfolio. Let
$f_\omega=\max_i|\{s\in\mathcal F:\max P_s=i\}|$. The algorithm runs in
$O(n2^\omega(1+f_\omega))$ time. Rolling value computation uses
$O(2^\omega)$ memory, while retaining reconstruction information uses
$O(n2^\omega)$ memory.

\subsection{Machine Layout}

\paragraph{Problem and global coupling.}
A feasible layout is a bijection $\pi$ from $n$ machines to $n$ linearly
ordered, unit-spaced slots, with $d_{st}=|s-t|$. Let $F_{ab}=F_{ba}$ be the
throughput between machines $a$ and $b$, and let
$A_{a,s}$ be the adjustment for placing machine $a$ in slot $s$. The objective
is
\begin{equation}
  \min_\pi\
  \sum_{a<b}F_{ab}d_{\pi(a),\pi(b)}
  +\sum_a A_{a,\pi(a)}.
\end{equation}
The cost of placing one machine depends on its distance from every other
machine with which it exchanges flow. The placement decisions are therefore
coupled through the complete permutation rather than through independent
machine--slot scores.

\paragraph{Exact algorithm and correctness.}
For a subset $S$ of machines, define
\begin{equation}
  \operatorname{cut}(S)=
  \sum_{a\in S}\sum_{b\notin S}F_{ab}.
\end{equation}
For a left-to-right order with prefix sets $S_t$, the pairwise flow--distance
cost is $\sum_{t=1}^{n-1}\operatorname{cut}(S_t)$. Because
$d_{st}=|s-t|$, a machine pair appears in one prefix cut for each unit slot
boundary between its two positions. Summing the cuts therefore reproduces the
complete flow--distance cost. With
$\operatorname{dp}[\varnothing]=0$, the recurrence
\begin{equation}
  \operatorname{dp}[S]=
  \min_{a\in S}
  \left\{
    \operatorname{dp}[S\setminus\{a\}]
    + A_{a,|S|}
    + \mathds{1}[|S|<n]\operatorname{cut}(S)
  \right\}
  \label{eq:layout-dp}
\end{equation}
conditions on the machine $a$ placed in slot $|S|$. The machines in
$S\setminus\{a\}$ must optimally fill the preceding slots. Each prefix cut and
placement adjustment is added exactly once, so induction on $|S|$ proves the
recurrence. The optimal value is
$\operatorname{dp}[\{1,\ldots,n\}]$, and stored minimizing machines
reconstruct the layout in reverse order. All cut values can be precomputed by
extending subsets one machine at a time. This takes $O(n2^n)$ time and
$O(2^n)$ space, after which each candidate in \cref{eq:layout-dp} takes
constant time. The algorithm therefore runs in $O(n2^n)$ time and uses
$O(2^n+n^2)$ memory.

\subsection{Sequential Matching}

\paragraph{Problem and global coupling.}
Let $L=(1,\ldots,n)$ be the fixed order of the left participants, let $R$ be
a set of $n$ right participants, and let
$E\subseteq L\times R$ contain the compatible pairs. A feasible decision is a
permutation $\pi=(\pi_1,\ldots,\pi_n)$ of $R$ satisfying
$(i,\pi_i)\in E$ for every position $i$. Each compatible pair $(i,j)\in E$
has unary match cost $c(i,j)$, and every ordered pair of distinct right
participants $(j,t)$ has directed handoff cost $h(j,t)$. The objective is
\begin{equation}
  \min_\pi\
  \sum_{i=1}^{n}c(i,\pi_i)
  +\sum_{i=2}^{n}h(\pi_{i-1},\pi_i).
\end{equation}
The bijection constraint couples assignments through the right participants
already used, while each handoff cost couples two consecutive matches.

\paragraph{Exact algorithm and correctness.}
For $S\subseteq R$ and $j\in S$, let $D[S,j]$ be the minimum cost of matching
the first $|S|$ left participants compatibly to exactly the right participants
in $S$, with $j$ assigned at position $|S|$. The initial states are
\begin{equation}
  D[\{j\},j]=c(1,j)
  \qquad\text{for every }(1,j)\in E,
\end{equation}
and all other states have value $+\infty$. For $t\in R\setminus S$ satisfying
$(|S|+1,t)\in E$,
\begin{equation}
  D[S\cup\{t\},t]
  =
  c(|S|+1,t)
  +\min_{j\in S}\{D[S,j]+h(j,t)\}.
  \label{eq:matching-dp}
\end{equation}
The optimal value is $\min_{j\in R}D[R,j]$. Every finite state represents a
compatible prefix with a specified set of right participants and a specified
final participant. The recurrence considers every compatible extension of
that prefix. Induction on $|S|$ therefore proves that each state retains the
cheapest compatible prefix with the same assigned set and final participant.
Stored predecessors reconstruct the matching.

There are at most $n2^{n-1}$ states $(S,j)$, and the running time is
$O(n^22^n)$.

\subsection{Fleet Dispatch}

\paragraph{Problem and global coupling.}
Let $J=\{1,\ldots,n\}$ be the jobs, let $V=\{1,\ldots,m\}$ be the vehicles,
and let $E\subseteq V\times J$ contain the eligible vehicle--job pairs. Job
$j$ has positive integer load $\lambda_j$. Vehicle $v$ has integer capacity
$C_v\geq0$, fixed activation cost $f_v$, and route cost $r_{vj}$ for each
$(v,j)\in E$. Binary variable $x_{vj}$ records an assignment. Vehicle
activation is determined by whether it receives at least one job. The
objective includes a decision-independent instance offset
$\gamma_{\mathcal I}$:
\begin{equation}
  g_{\mathcal I}(x)
  =\gamma_{\mathcal I}
  +
  \sum_{(v,j)\in E}r_{vj}x_{vj}
  +\sum_{v\in V}f_v
    \mathds{1}\!\left[\sum_{j:(v,j)\in E}x_{vj}>0\right].
\end{equation}
The problem minimizes $g_{\mathcal I}(x)$ subject to
\begin{equation}
\begin{aligned}
  &\sum_{v:(v,j)\in E}x_{vj}=1
    &&\forall j\in J,\\
  &\sum_{j:(v,j)\in E}\lambda_jx_{vj}\leq C_v
    &&\forall v\in V,\\
  &x_{vj}\in\{0,1\}
    &&\forall (v,j)\in E.
\end{aligned}
\end{equation}
Jobs assigned to the same vehicle consume a shared capacity and share one
activation cost. Assigning one job therefore changes both the remaining
capacity and the marginal cost of later assignments to that vehicle.

\paragraph{Exact algorithm and correctness.}
Process jobs in their fixed order. Let
$\boldsymbol{\ell}=(\ell_1,\ldots,\ell_m)$ be the assigned vehicle loads and
let $D_i(\boldsymbol{\ell})$ be the minimum cost of assigning the first $i$
jobs with that load vector, excluding the decision-independent offset
$\gamma_{\mathcal I}$. The vector is encoded as one mixed-radix index with
radices $C_v+1$. Initialize $D_0(\boldsymbol 0)=0$ and every other state to
$+\infty$. For every reached state $\boldsymbol\ell$ and every
$(v,i+1)\in E$ satisfying $\ell_v+\lambda_{i+1}\leq C_v$, update
\begin{equation}
\begin{split}
  D_{i+1}\!\left(\boldsymbol{\ell}+\lambda_{i+1}\mathbf e_v\right)
  \leftarrow\min\Bigl\{&
  D_{i+1}\!\left(\boldsymbol{\ell}+\lambda_{i+1}\mathbf e_v\right),\\
  &D_i(\boldsymbol{\ell})+r_{v,i+1}
  +f_v\mathds{1}[\ell_v=0]\Bigr\}.
\end{split}
\label{eq:fleet-dp}
\end{equation}
Because loads are positive, $\ell_v=0$ holds exactly when $v$ has not yet
been activated. Given the next job index, the load vector determines all
remaining capacities and which activation costs have already been incurred.
No other property of the prefix affects a continuation. Every feasible
assignment of the first $i+1$ jobs assigns job $i+1$ to exactly one vehicle, so
the recurrence considers every extension at this step. Induction over the
job order proves the recurrence. The optimal objective value is
$\gamma_{\mathcal I}+\min_{\boldsymbol{\ell}}D_n(\boldsymbol{\ell})$, and
stored predecessors reconstruct the job-to-vehicle mapping.

Writing $P=\prod_{v=1}^{m}(C_v+1)$, the algorithm runs in $O(nmP)$ time and
uses $O(nP+nm)$ memory when storing reconstruction information.

\subsection{Evidence-Joined Routing}

\paragraph{Problem and global coupling.}
The instance has a positive residue modulus $M$, a target residue
$R\in\{0,\ldots,M-1\}$, and a layered directed acyclic graph $G=(V,E)$ with
source $s$ in the first layer and target $t$ in the last. Every edge
points from an earlier layer to a later layer. Each edge $e$ has quantity
$Q_e$, exception class $\chi_e$, residue
$\rho_e\in\{0,\ldots,M-1\}$, and tariff $T(\chi_e)$. The quantity and
exception-class tariff define the nonnegative edge cost
$\kappa_e=Q_eT(\chi_e)$. A feasible decision is an $s$--$t$ path $P$ whose
accumulated residue satisfies
\begin{equation}
  \sum_{e\in P}\rho_e\equiv R\pmod M,
\end{equation}
and the objective is to minimize $\sum_{e\in P}\kappa_e$. Edge costs combine
quantity, exception-class, and tariff data, while path feasibility depends on
the residue accumulated across all selected edges.

\paragraph{Exact algorithm and correctness.}
Let $D(v,r)$ be the minimum cost of a path from $s$ to $v$ with residue $r$.
Initialize $D(s,0)=0$ and all other states to $+\infty$. Traversing an
edge $e=(u,v)$ induces the expanded-state relaxation
\begin{equation}
\begin{split}
  D\!\left(v,(r+\rho_e)\bmod M\right)
  \leftarrow\min\Bigl\{&D\!\left(v,(r+\rho_e)\bmod M\right),\\
               &D(u,r)+\kappa_e\Bigr\}.
\end{split}
  \label{eq:evidence-routing-transition}
\end{equation}
Processing the layered graph in topological order therefore computes
$D(t,R)$, and stored predecessors reconstruct the path.
The state records exactly the endpoint and accumulated residue on which a
continuation depends, so induction over the graph layers establishes the
recurrence's correctness.

Writing $n_V=|V|$ and $n_E=|E|$ for the numbers of vertices and edges in the
graph, the algorithm runs in $O(M(n_V+n_E))$ time.

\subsection{Migration Portfolio}

\paragraph{Problem and global coupling.}
Let $\mathcal D=\{1,\ldots,n_D\}$ be the support domains, and let
$\mathcal P_i$ be the package catalog for component group
$i\in\{1,\ldots,G\}$. Package $p\in\mathcal P_i$ has cost $c_p$ and
requirement set $R(p)\subseteq\mathcal D$. A decision
$y=(S,p_1,\ldots,p_G)$ activates a domain set $S\subseteq\mathcal D$ and
selects one package $p_i\in\mathcal P_i$ per group. It is feasible when
$R(p_i)\subseteq S$ for every group. Activating domain $j$ incurs activation
cost $F_j$.

The instance also has a finite set $\mathcal H$ of signed interaction factors.
Each $a\in\mathcal H$ has a nonempty domain scope $P_a\subseteq\mathcal D$
and a table $\phi_a:\{0,1\}^{|P_a|}\rightarrow\mathbb Z$. Define
\begin{equation}
  \Phi(S)=
  \sum_{a\in\mathcal H}
  \phi_a\!\left(\mathds{1}_S|_{P_a}\right),
\end{equation}
and let $L$ be a fixed instance-specific constant. The term $-L+1$ is
independent of the decision and does not change the minimizing decision. The
objective of a feasible decision is
\begin{equation}
  \min_y\
  \left\{
    \sum_{j\in S}F_j+
    \sum_{i=1}^{G}c_{p_i}+
    \Phi(S)-L+1
  \right\}.
\end{equation}
Package choices are coupled because they can require the same activated
domains, while the signed factors depend jointly on several domain statuses.

\paragraph{Exact algorithm and correctness.}
For $S\subseteq\mathcal D$, let
\begin{equation}
  f_i(S)=
  \min_{\substack{p\in\mathcal P_i\\R(p)\subseteq S}}c_p,
\end{equation}
with $f_i(S)=+\infty$ when no package in group $i$ is compatible with $S$.
The construction algorithm evaluates every domain mask using
\begin{equation}
  V(S)=
  \sum_{j\in S}F_j+
  \sum_{i=1}^{G}f_i(S)+
  \Phi(S)-L+1.
  \label{eq:migration-mask-objective}
\end{equation}
For any fixed $S$, the package choices separate across groups, so the stored
minimizer for each finite $f_i(S)$ yields a feasible decision whose objective
is exactly $V(S)$. Conversely, every feasible decision with activated set $S$
has package cost at least $\sum_i f_i(S)$. Minimizing $V(S)$ over all masks
therefore gives the global optimum; the stored groupwise minimizers reconstruct
the complete decision.

The construction algorithm uses the least-significant set bit to update the
activation-cost and interaction contributions incrementally. It scans the
package catalogs to obtain the groupwise minima. Define the per-mask work bound
\begin{equation}
  I=n_D+G+
  \sum_{i=1}^{G}\sum_{p\in\mathcal P_i}\bigl(1+|R(p)|\bigr)+
  \sum_{a\in\mathcal H}\bigl(1+|P_a|\bigr).
\end{equation}
The algorithm runs in
$O(I2^{n_D})$ time and uses $O(2^{n_D}+G)$ additional working memory.

\section{Deterministic Instance Generation}
\label{app:generation}

Section~\ref{sec:synthesis} describes instance generation using a size
configuration, a structure configuration, and a deterministic generator. For
each task family and workload level, we fix one size configuration and one
structure configuration. These configurations, together with an instance
index $j\in\{1,2,3\}$, determine the input to the family's human-written
generator and thus one hidden instance $\mathcal I$. Repeating the construction
with the same family, workload level, and instance index reproduces the same
entities, relations, coefficients, and constraints.

\paragraph{Size and structure controls.}
The size configuration governs scale-bearing quantities such as the numbers
of entities, choices, stages, or relations. The structure configuration
governs the relations through which decisions interact in the constraints and
objective. A single quantity can affect both problem scale and the pattern of
dependencies among decisions, so the two roles need not correspond to separate
generator arguments. \Cref{tab:generation-controls} summarizes the principal
quantities and coupling mechanisms for each family. The table is not a complete
list of generator arguments.

\begin{table}[H]
\centering
\begingroup
\awtablecompact
\awtablezebra{2}
\resizebox{0.95\textwidth}{!}{%
\begin{tabular}{@{}
  >{\raggedright\arraybackslash}p{3.0cm}
  >{\raggedright\arraybackslash}p{4.5cm}
  >{\raggedright\arraybackslash}p{6.5cm}@{}}
\toprule
\awtableheader
Task family & Principal scale-bearing quantities & Structure producing global
coupling \\
\midrule
Transit Routing & Numbers of stages, candidate legs, and lines & Residue
condition and directed transfer costs between consecutive lines \\
Basket Assembly & Item-group and ticket-catalog sizes & Ticket--item
incidence and exact-cover overlap \\
Station Siting & Numbers of zones and demand relations & Shared site-count
constraint and bounded-span pairwise demand terms \\
Authorization Planning & Numbers of stages, packages, and authorized
transitions & Authorized transitions, handoffs, and the global clearance
residue \\
Series Portfolio & Numbers of titles, lots, and factors & Overlapping bounded
factor scopes \\
Machine Layout & Numbers of machines and slots & Permutation feasibility and
pairwise flow terms \\
Sequential Matching & Participant and compatible-pair counts & Bijective
matching and handoff costs between consecutive matches \\
Fleet Dispatch & Numbers of jobs and vehicles & Eligibility, competition for
vehicle capacity, and shared activation costs \\
Evidence-Joined Routing & Layer, vertex, and edge counts & Joined edge
attributes, path continuity, and the global residue condition \\
Migration Portfolio & Numbers of component groups, packages, and support
domains & Shared domain activation and signed cross-domain interactions \\
\bottomrule
\end{tabular}%
}
\caption{Scale-bearing quantities and global-coupling structures represented
by the ten formal problem families. The entries summarize semantic roles in
the formal instances rather than a complete list of generator arguments; the
exact algorithms and their complexity parameters are specified in
\cref{app:cases,app:work}.}
\label{tab:generation-controls}
\endgroup
\end{table}

\paragraph{Fixed membership and reproducibility.}
The family, workload level, and instance index are fixed before model
evaluation. We validate the instance at each fixed family--level--index
coordinate rather than selecting a replacement based on validation results or
model outcomes. The released benchmark contains three fixed instances for each
family--level combination, and all of them pass the required checks. The serialized hidden
instances record this benchmark membership. The manuscript describes the
construction contract, formal problems, and validation checks used for these
instances.

\section{Tool-Mediated World Construction}
\label{app:world-construction}

\paragraph{Reusable family templates.}
World construction turns a generated hidden instance $\mathcal I$ into an
environment that an agent can access through tools. For each family, a reusable
template instantiates the textual question, maps fields of $\mathcal I$ to
interface records, defines the tool schemas, assigns tool-call costs and an
access budget, and specifies the final-decision format.
\Cref{tab:world-template-components} distinguishes the components shown to the
agent from the hidden instance and private records used to produce tool
responses.

\begin{table}[H]
\centering
\begingroup
\awtablecompact
\awtablezebra{2}
\resizebox{0.95\textwidth}{!}{%
\begin{tabular}{@{}
  >{\raggedright\arraybackslash}p{3.1cm}
  >{\raggedright\arraybackslash}p{5.1cm}
  >{\raggedright\arraybackslash}p{6.0cm}@{}}
\toprule
\awtableheader
Construction artifact & Template action & Paired-world relation and agent
visibility \\
\midrule
Hidden instance $\mathcal I$ & Supplies the constraints, objective terms, and
case-specific values & Shared by the paired worlds and not provided directly
to the agent. \\
Textual question $q$ & Instantiates the family task statement and the public
identifiers needed to describe and submit a decision & Identical in the paired
worlds and provided to the agent. \\
Private interface records & Maps the fields of $\mathcal I$ into the
relational records queried by an interface's tools & Organized differently in
the Direct and Mediated worlds; selected records appear only through executed
tool responses. \\
Information tools $\mathcal T_{\mathrm{info}}$ & Defines the callable
operations, argument schemas, and record queries & The agent receives the tool
schemas and the responses to executed calls, but not the complete private
record collection. \\
Cost function $f_c$ and budget $b$ & Assigns per-call costs and the total
amount available for one trajectory & Identical in the paired worlds and
provided to the agent. \\
Final-decision tool $t_{\mathrm{dec}}$ & Defines the family's structured
decision format and records the submitted decision & Identical in the paired
worlds; its schema is provided to the agent, and an executed call commits the
final decision. \\
\bottomrule
\end{tabular}%
}
\caption{Components used or instantiated by a family-specific world-construction
template. Paired worlds share the decision problem and agent-facing task
conditions while using different private relational organizations for the
instance information.}
\label{tab:world-template-components}
\endgroup
\end{table}

\paragraph{Paired relational records.}
Applying the two interface mappings to the same $\mathcal I$ produces private
Direct and Mediated record collections, denoted
$R_{\mathcal I}^{(\mathrm D)}$ and $R_{\mathcal I}^{(\mathrm M)}$,
respectively. These collections encode the same instance rather than adding
new optimization data. A collection may introduce interface-specific
identifiers or distribute one decision-relevant relation across several
linked records. The information tools return only the records selected by the
calls made along a trajectory. The paired worlds retain the same $\mathcal I$,
$q$, $f_c$, $b$, and final-decision format. When a candidate-evaluation tool is
included, its schema, cost, and checker behavior are also shared by the pair.
The checks in
\cref{app:paired-equivalence} confirm that both private record collections
recover the same hidden instance, although their raw records and tool responses
need not match.

\paragraph{Candidate evaluation.}
Some worlds additionally include a candidate-evaluation tool
$t_{\mathrm{eval}}\in\mathcal T_{\mathrm{info}}$. It accepts one complete
candidate decision and reports whether that decision is feasible and, when it
is feasible, its objective value. The tool has positive cost, is nonterminal,
and may be invoked at most once. It neither commits the candidate as the final
decision nor directly returns fields of $\mathcal I$; only a later call to
$t_{\mathrm{dec}}$ submits a binding final decision. This tool-specific
one-call limit applies even though the other information tools may be invoked
repeatedly within the trajectory limits.

Its feedback appears in the recorded trajectory and contributes to tool
expenditure, but it does not add instance information to the trajectory
diagnostics. The corresponding accounting appears in
\cref{app:acquisition,app:metrics}.

\paragraph{Released specifications.}
For each world, the instantiated specification fixes the textual question,
tool schemas, mappings between calls and private records, tool-call costs,
access budget, and final-decision format. The offline validation records also
contain the two information-gathering call lists described in
\cref{sec:data-quality}. Each list specifies the ordered information-tool calls
and their arguments, and its total cost follows from $f_c$.

\Cref{app:acquisition} defines the request form, cost, response normalization,
sufficiency criterion, and budget conditions for these lists.
\Cref{app:worked-world-validation} presents both lists and their costs for one
released Fleet Dispatch world. The 480 instantiated lists are retained in the released
validation records rather than printed in the manuscript, and they are not
provided to the evaluated agents.

\section{Offline Validation}
\label{app:certificates}

Before model evaluation, we validate the recorded optima, final-decision
scoring, information access, and paired-interface construction described in
\cref{sec:synthesis,sec:data-quality}. We also verify the integrity of the
released world bundles. Only worlds that pass all applicable checks are
included in the benchmark. Because the checks use different validation units,
the summary at the end of the section reports each unit separately.

\subsection{Instance and Scoring Checks}

\paragraph{Why independent verification is required.}
The recurrences and correctness arguments in \cref{app:cases} establish that
the construction algorithms are exact under the formal problem definitions.
The recorded optimum, however, is produced by a concrete implementation of the
state encoding, objective accounting, and decision reconstruction. We therefore
solve every hidden instance again with an independently implemented exact
algorithm that uses a different state direction or problem representation.
Agreement between the two algorithms provides an independent check of the value
and decision recorded during construction.

\paragraph{Independent optima.}
\Cref{tab:verification-methods} summarizes how the independent verification
algorithm
differs from the construction algorithm described for each family in
\cref{app:cases}. A unique optimum means that exactly one feasible structured
decision attains the optimal objective value. A fixed instance passes this
check only when the two implementations agree on that decision and value and
both confirm uniqueness. All fixed instances pass.

\begin{table}[H]
\centering
\begingroup
\awtablecompact
\awtablezebra{2}
\resizebox{0.95\textwidth}{!}{%
\begin{tabular}{@{}
  >{\raggedright\arraybackslash}p{3.1cm}
  >{\raggedright\arraybackslash}p{11.0cm}@{}}
\toprule
\awtableheader
Task family & Independent verification algorithm \\
\midrule
Transit Routing & Represents the computation as exhaustive relaxation over the
dense stage--line--residue graph rather than as the construction run's ordered
predecessor scan. \\
Basket Assembly & Uses a forward-mask recurrence over the tripartite coverage
state rather than the first-uncovered-item recurrence. \\
Station Siting & Uses a top-down recurrence with the generic at-most-$H$
capacity semantics rather than the construction run's saturation reduction.
\\
Authorization Planning & Uses a reverse suffix recurrence rather than the
forward prefix recurrence. \\
Series Portfolio & Scans from right to left and closes each factor at its
leftmost position rather than scanning left to right and closing it at its
rightmost position. \\
Machine Layout & Builds the layout from the first slot by choosing the next
machine rather than choosing the final machine of each prefix. \\
Sequential Matching & Processes suffixes in reverse and conditions on their
first right participant rather than processing prefixes and retaining their
last match. \\
Fleet Dispatch & Uses remaining capacities in a top-down recurrence rather
than assigned loads in a forward recurrence. \\
Evidence-Joined Routing & Represents the computation as a general nonnegative
shortest-path problem on the expanded state graph and runs Dijkstra's
algorithm rather than the construction run's topological dynamic program. \\
Migration Portfolio & Represents consecutive domain masks in Gray-code order,
updates the single changed domain, and independently recomputes compatible
groupwise package minima. \\
\bottomrule
\end{tabular}%
}
\caption{Independent verification algorithms used to validate the optima produced by
the ten construction algorithms.}
\label{tab:verification-methods}
\endgroup
\end{table}

The work record comes only from the construction algorithm, as
described in \cref{app:work}; the independent verification algorithm serves only to
validate its output. All 120 hidden instances pass the agreement and
uniqueness checks.

\paragraph{Decision checking.}
For each of the 240 worlds, the family-specific decision checker determines
whether a structured decision $y$ satisfies the family's feasibility
conditions. When it does, the checker recomputes the objective value directly
from $\mathcal I$. As a consistency check, it must accept the independently
verified optimal decision and reproduce the verified value $v_w^\star$.

\paragraph{Reference decisions.}
Let $y_{\mathrm{ref}}(w)$ denote the suboptimal reference decision used to set
the reference-utility scale for world $w$. Before model evaluation, the checker
accepts this decision as feasible and verifies
\begin{equation}
  F_w\!\left(y_{\mathrm{ref}}(w)\right)=1,
  \qquad
  \Delta_w^{\mathrm{ref}}
  =\Delta_w\!\left(y_{\mathrm{ref}}(w)\right)>0.
  \label{eq:reference-validation}
\end{equation}
Thus the denominator in \cref{eq:reference-utility} is positive. Because the
objectives are integer-valued, exact optimality is checked by equality without
a floating-point tolerance. All 240 worlds pass the optimal-decision and
reference-decision checks. The fixed scoring record for a world stores the
verified optimal decision and value together with the validated reference
decision and its recomputed gap.

\subsection{Acquisition and Information-Separation Checks}
\label{app:acquisition}

\paragraph{From information tools to instance facts.}
The main text uses a \emph{sufficient information set} to mean instance
information that is enough to reconstruct $\mathcal I$ and determine its
verified optimum. For formal comparison across interfaces, we represent this
information as \emph{normalized instance facts}: canonical facts about
$\mathcal I$ obtained after combining and interpreting the relevant tool
responses.

For a world $w$, let $\mathcal T_{\mathrm{info}}(w)$ denote its
information-tool suite. We call a tool \emph{fact-revealing} when its responses
expose normalized instance facts, and denote the set of these tools by
$\mathcal T_{\mathrm{fact}}(w)\subseteq\mathcal T_{\mathrm{info}}(w)$. The
candidate-evaluation tool defined in \cref{app:world-construction} belongs to
$\mathcal T_{\mathrm{info}}(w)\setminus\mathcal T_{\mathrm{fact}}(w)$ because
it returns feedback on a proposed decision rather than fields of
$\mathcal I$. The final-decision tool $t_{\mathrm{dec}}$ is likewise not a
fact-revealing tool.

\paragraph{Acquisition plans.}
\Cref{sec:data-quality} describes two alternative call lists. Each specifies
the fact-revealing tools and their arguments. We formalize each such list as an
\emph{acquisition plan}. A fact-revealing request is a
structured call $a=(t,x)$ to a fact-revealing tool with an admissible argument
tuple. An acquisition plan for $w$ is a finite
ordered sequence $P=(a_1,\ldots,a_m)$ of these requests, with
$a_i=(t_i,x_i)$. Its cost is
\begin{equation}
  \operatorname{cost}_w(P)=\sum_{i=1}^{m}f_c(t_i).
  \label{eq:plan-cost}
\end{equation}
Neither the candidate-evaluation tool nor the final-decision tool is part of an
acquisition plan. A plan is an offline validation object, not a trajectory or
an acquisition strategy prescribed to an evaluated agent.

\paragraph{Normalizing combined responses.}
Raw Direct and Mediated responses may organize the same instance information
through different relational records, and one fact may require records from
more than one response. We therefore compare their interface-neutral content
rather than their raw fields or tokens. When request $a_i$ returns response
$o_i$, a plan produces the call--response record sequence
\[
  ((a_1,o_1),\ldots,(a_m,o_m)).
\]
The world-specific response normalizer $\operatorname{NF}_w$ is defined on any
finite ordered sequence of executed fact-revealing call--response records. It
parses the returned records and performs the joins required across responses.
It then removes interface-only identifiers and records, then deduplicates the
resulting normalized instance facts. Applying it to the sequence produced by
$P$ gives
\begin{equation}
  S_w(P)=\operatorname{NF}_w
  \bigl((a_1,o_1),\ldots,(a_m,o_m)\bigr).
  \label{eq:plan-facts}
\end{equation}

\paragraph{Sufficient information sets.}
A plan is sufficient when its nonempty set $S_w(P)$ reconstructs $\mathcal I$
and the family-specific construction algorithm applied to that reconstruction recovers
the same independently verified optimum. Thus $S_w(P)$ is the formal
instance-fact representation of a \emph{sufficient information set} in
\cref{sec:metrics}.

Each world has two validated acquisition plans,
\begin{equation}
  \mathcal P(w)=\{P_w^{(1)},P_w^{(2)}\},
  \qquad
  \mathcal S(w)=\{S_w(P):P\in\mathcal P(w)\}.
  \label{eq:plan-set-family}
\end{equation}
The two plans may produce the same normalized fact set, so
$|\mathcal S(w)|$ may be one or two. Both cases are covered by the trajectory
diagnostics in \cref{sec:metrics}. All 480 plans execute successfully and are
sufficient. These sets need not be necessary or minimal, and the two plans do
not enumerate every sufficient subset or prescribe an acquisition strategy to
the evaluated agent.

\paragraph{Budget conditions.}
For every fact-revealing tool $t$, let $\mathcal X_w(t)$ be its declared finite
set of admissible argument tuples. The complete catalog of requests that can
return instance information, and the cost of issuing each distinct request
once, are
\begin{equation}
  \mathcal C_{\mathrm{fact}}(w)
  =\{(t,x):t\in\mathcal T_{\mathrm{fact}}(w),\ x\in\mathcal X_w(t)\},
  \qquad
  c_{\mathrm{all}}(w)
  =\sum_{(t,x)\in\mathcal C_{\mathrm{fact}}(w)}f_c(t).
  \label{eq:fact-catalog}
\end{equation}
Each distinct admissible request appears once; invalid calls, repeated copies,
$t_{\mathrm{eval}}$, and $t_{\mathrm{dec}}$ are excluded. Let
\begin{equation}
  c_{\mathrm{eval}}(w)=
  \begin{cases}
    f_c(t_{\mathrm{eval}}), & t_{\mathrm{eval}}\text{ is available in }w,\\
    0, & \text{otherwise}.
  \end{cases}
\end{equation}
Validation requires
\begin{equation}
\begin{aligned}
  \operatorname{cost}_w(P)+c_{\mathrm{eval}}(w)&\leq b
    &&\forall P\in\mathcal P(w),\\
  |P|+\mathds{1}[t_{\mathrm{eval}}\text{ is available in }w]&\leq255
    &&\forall P\in\mathcal P(w),\\
  b&<c_{\mathrm{all}}(w).
\end{aligned}
  \label{eq:access-envelope}
\end{equation}
The first two inequalities leave enough budget and calls to execute a validated
plan and, where available, one candidate-evaluation call. The reserve
$c_{\mathrm{eval}}(w)$ is charged only if the agent invokes
$t_{\mathrm{eval}}$. The final inequality states that the budget does not
permit issuing every request in the catalog. Together, the conditions show that
a sufficient set of calls fits within both trajectory limits whereas the full
catalog does not fit within the access budget. All 480 plans satisfy the first
two conditions, and all 240 worlds satisfy the full-catalog condition.

\paragraph{Information-separation checks.}
These checks examine whether decision-relevant information is distributed
across the tool interface rather than concentrated in one channel or response.
We group normalized instance facts by the kind of information they describe
and call each group an \emph{information channel}. A channel is a validation
grouping, not necessarily one tool, field, source record, or response. Every
validated acquisition plan contains facts from multiple channels.

For each world--channel pair, we construct a counterfactual instance under the
family's formal definition. Observations from all other channels remain fixed,
while information in the tested channel changes so that the verified optimal
decision differs. All 648 world--channel checks pass. Each tested channel can therefore affect the
optimal decision under the corresponding counterfactual. This does not imply
that every fact in the channel is necessary or that the channel is required to
reconstruct every instance.

The response-level check constructs a counterfactual instance that holds one
selected information-tool response fixed while modifying an unrevealed
normalized instance fact so that the verified optimal decision changes. All
1,464 selected-response checks pass. None of the
tested responses alone therefore determines the optimum. This finite result
does not extend to every possible request, argument, or response. Neither the
separation checks nor the access envelope establishes an information-theoretic
lower bound.

\paragraph{Information and access scale.}
\begin{table}[H]
\centering
\begingroup
\awtablestandard
\setlength{\tabcolsep}{2.0pt}
\awtablezebra{2}
\resizebox{1.00\textwidth}{!}{%
\begin{tabular}{@{}>{\raggedright\arraybackslash}p{2.8cm}*{6}{c}@{}}
\toprule
\awtableheader
\textbf{Task family} & \makecell{\textbf{Info.}\\\textbf{groups}} & \makecell{\textbf{Sufficient}\\\textbf{set size}} & \makecell{\textbf{Call}\\\textbf{count}} & \makecell{\textbf{Call}\\\textbf{cost}} & \makecell{\textbf{Budget}\\\textbf{/ exhaustive}} & \makecell{\textbf{Algorithmic}\\\textbf{work}} \\
\midrule
Transit Routing & 3 & 1,755--6,963 & 59--147 & 226--634 & 524--884 / 637--1,558 & 50,737--47,619,285 \\
Basket Assembly & 2 & 262--640 & 38--180 & 172--764 & 260--780 / 1,020--2,554 & 35,646--39,098,995 \\
Station Siting & 2 & 284--326 & 172--234 & 558--774 & 577--779 / 915--1,293 & 55,391--34,318,974 \\
Authorization Planning & 3 & 2,633--7,781 & 17--73 & 99--291 & 120--296 / 729--1,257 & 43,747--35,313,268 \\
Series Portfolio & 2 & 381--1,458 & 24--141 & 196--924 & 1,228--1,924 / 1,491--5,658 & 54,587--34,402,415 \\
Machine Layout & 3 & 444--1,300 & 10--42 & 93--370 & 129--377 / 963--3,166 & 111,668--46,226,237 \\
Sequential Matching & 3 & 511--1,463 & 22--52 & 122--308 & 175--313 / 582--1,447 & 107,654--39,865,720 \\
Fleet Dispatch & 3 & 183--405 & 11--39 & 51--175 & 97--180 / 288--705 & 42,042--53,964,927 \\
Evidence-Joined Routing & 3 & 2,166--18,150 & 11--51 & 72--202 & 343--452 / 428--1,092 & 88,935--52,827,071 \\
Migration Portfolio & 3 & 707--1,829 & 11--151 & 252--1,542 & 277--1,547 / 2,142--11,344 & 61,393--46,070,264 \\
\bottomrule
\end{tabular}%
}
\caption{Information and access scale by task family. Sufficient-information-set
size, call count, and call cost range over 48 world--call-list combinations
(two validated call lists for each of 24 worlds); call cost excludes the
candidate-feedback reserve. Budget and exhaustive-catalog cost are separate
world-level ranges. Algorithmic
work ranges over the 12 underlying hidden instances, whose paired worlds
inherit the same value defined in \cref{eq:algorithm-work}.}
\label{tab:benchmark-statistics}
\endgroup
\end{table}

\Cref{tab:benchmark-statistics} reports information channels, sufficient-set
sizes, call counts, and call costs alongside algorithmic work. For each family,
the sufficient-set, call-count, and call-cost ranges cover the two validated
acquisition plans for each of its 24 worlds. Set size counts normalized instance
facts after the relevant responses have been combined. Call cost excludes the
candidate-evaluation reserve included in \cref{eq:access-envelope}.

Budget and full-catalog cost are separate world-level ranges over the 24
worlds. Algorithmic work is summarized over the 12 underlying hidden instances
and shared by each Direct--Mediated pair. The columns use different scales and
aggregation units. Their range endpoints need not describe the same world and
cannot be combined to infer a within-world access guarantee.

\subsection{Paired-Interface Equivalence}
\label{app:paired-equivalence}

Recall from \cref{app:world-construction} that
$R_{\mathcal I}^{(\mathrm D)}$ and $R_{\mathcal I}^{(\mathrm M)}$ are the
complete private record collections used by a Direct--Mediated pair. Let
$N^{(\mathrm D)}$ and $N^{(\mathrm M)}$ be the interface-specific mappings
that reconstruct an instance from these complete record collections. Pair
validation requires
\begin{equation}
  N^{(\mathrm D)}\!\left(R_{\mathcal I}^{(\mathrm D)}\right)
  =\mathcal I
  =N^{(\mathrm M)}\!\left(R_{\mathcal I}^{(\mathrm M)}\right).
  \label{eq:paired-normalization}
\end{equation}
All 120 pairs pass this equality. The common hidden instance recovered by
normalization defines the same feasible decisions and objective function, and
both worlds inherit its verified optimum, uniqueness result, and algorithmic-
work record. A separate structural check confirms that the interface-specific
relational records differ before normalization.

The pair-level mappings $N^{(\mathrm D)}$ and $N^{(\mathrm M)}$ consume the
complete private record collections. They are distinct from
$\operatorname{NF}_w$, which combines only the responses returned by the
information-tool calls in a particular plan or trajectory. This distinction
allows a pair to encode the same complete instance even when an agent observes
only different subsets of its records.

As specified in \cref{sec:synthesis}, the paired worlds also retain the same
textual question, tool-call costs, access budget, final-decision format, and,
when included, candidate-evaluation tool.
The equivalence check does not require identical raw responses or tool-use
trajectories, nor does it imply identical model performance. Its conclusions
apply only to the validated interface pairs.

\subsection{Release Integrity}
\label{app:release-integrity}

The release manifest binds each of the 240 world bundles to its serialized
hidden instance, textual question, tool schemas, scoring record, validation
records, and supporting documentation through content digests. A missing,
altered, extra, or mismatched component causes the integrity check to fail.
Only bundles that pass all applicable validation checks above are included in
the released benchmark.

The verified optimum and uniqueness result, work record from the construction algorithm,
validated acquisition plans, and sufficient information sets remain offline
benchmark records and are not supplied to an evaluated agent. Model evaluation
uses the fixed world and scoring records and does not rerun either exact
algorithm. Evaluation-record integrity is discussed separately in
\cref{app:operations}.

\subsection{Worked Fleet Dispatch Validation}
\label{app:worked-world-validation}

We now apply the preceding checks to the released pair introduced in
\cref{app:worked-world}. Its tools and per-call costs are listed in
\cref{tab:worked-world-tools}.

\paragraph{Two validated ways of gathering sufficient information.}
For the Direct world, the two call lists described in the main text become the
acquisition plans $P_{\mathrm{row}}$ and $P_{\mathrm{window}}$.
$P_{\mathrm{row}}$ uses one policy call, 14 job calls, five contract calls, and
five route-row calls; $P_{\mathrm{window}}$ uses one policy call, five contract
calls, and five window calls. The five window calls collectively
cover all 14 jobs; their responses supply the loads and eligibility relation
in addition to the route records. Their call counts and costs are
\begin{equation}
\begin{aligned}
  |P_{\mathrm{row}}|&=25,
  &\operatorname{cost}_{w^{(\mathrm D)}}(P_{\mathrm{row}})
    &=2+14(1)+5(4)+5(3)=51,\\
  |P_{\mathrm{window}}|&=11,
  &\operatorname{cost}_{w^{(\mathrm D)}}(P_{\mathrm{window}})
    &=2+5(4)+5(14)=92.
\end{aligned}
\label{eq:worked-plan-costs}
\end{equation}
Normalizing the combined responses returned by either plan yields a sufficient
information set that reconstructs $\mathcal I$ and recovers its verified
optimum. Thus
\begin{equation}
\begin{aligned}
  \mathcal P(w^{(\mathrm D)})
    &=\{P_{\mathrm{row}},P_{\mathrm{window}}\},\\
  \mathcal S(w^{(\mathrm D)})
    &=\{S_{w^{(\mathrm D)}}(P_{\mathrm{row}}),
       S_{w^{(\mathrm D)}}(P_{\mathrm{window}})\}.
\end{aligned}
\label{eq:worked-plan-sets}
\end{equation}
The two normalized information sets need not be distinct. The corresponding
Mediated plans have the same call compositions and costs. Their response
records pass through the interface-specific joins illustrated above before the
normalized instance facts are compared.

\paragraph{Budget arithmetic.}
The candidate-evaluation tool costs five units, so the two validated plans
satisfy
\begin{equation}
  51+5\leq97,
  \qquad
  92+5=97,
  \qquad
  97<401,
  \label{eq:worked-access-envelope}
\end{equation}
where 401 is the cost of the full fact-request catalog. This total
expands as
$2+14(1)+5(4)+5(3)+(13+12)(14)=401$: with 14 ordered jobs, there are 13
admissible length-two windows and 12 admissible length-three windows. The
five-unit reserve is not part of either plan and is charged only if the agent
invokes the candidate-evaluation tool. The plans and their sufficiency checks
are offline validation objects, not trajectories prescribed to the agent.

\paragraph{Offline validation results.}
The construction and independent verification algorithms agree that the hidden
instance has a unique optimum. Its objective value is 262,456,246,691, and the
construction algorithm records algorithmic work of 45,799 operations.
Because $w^{(\mathrm D)}$ and $w^{(\mathrm M)}$ share $\mathcal I$, both
inherit this instance-level optimum and work record. A feasible submitted
mapping is exactly optimal when its recomputed objective equals the verified
value. The fixed reference decision and its gap remain part of the scoring
record.

The verified optimal decision, uniqueness result, work record, validated
plans, and sufficient information sets are offline records. They are not
included in $q$, the tool schemas, or model-visible tool responses.
If invoked, the candidate-evaluation tool reveals only the checker result for
the mapping supplied by the agent; it does not disclose the verified optimum
as a reference value.

\subsection{Validation Summary}

The checks above operate on several different units. \Cref{tab:validation-summary}
reports each unit separately.

\begin{table}[H]
\centering
\begingroup
\awtablecompact
\awtablezebra{2}
\resizebox{1.00\textwidth}{!}{%
\begin{tabular}{@{}
  >{\raggedright\arraybackslash}p{3.0cm}
  >{\raggedright\arraybackslash}p{3.0cm}
  >{\centering\arraybackslash}p{1.3cm}
  >{\raggedright\arraybackslash}p{7.2cm}@{}}
\toprule
\awtableheader
Validated property & Validation unit & Count & Pass criterion \\
\midrule
Fixed construction membership & Hidden instance & 120 & Each predetermined
family--level--index coordinate yields its reproducible instance. \\
Optimum agreement and uniqueness & Hidden instance & 120 & The construction
and independent verification algorithms agree on the optimal decision and value and confirm a
unique optimum. \\
Decision checking & Algorithmic world & 240 & The checker accepts the verified
optimal decision and reproduces its objective value. \\
Reference decision & Algorithmic world & 240 & The reference is feasible and
has $\Delta_w^{\mathrm{ref}}>0$. \\
Sufficient information acquisition & Acquisition plan & 480 & The responses
reconstruct $\mathcal I$, recover its verified optimum, span multiple
information channels, and fit within budget with the candidate-evaluation
reserve and within the information-call cap. \\
Full-catalog budget & Algorithmic world & 240 & Issuing every request in the
finite fact-revealing catalog exceeds $b$. \\
Channel counterfactual & World--channel pair & 648 & Other channel observations
remain fixed while the verified optimal decision changes. \\
Response isolation & Selected response & 1,464 & The tested response remains
fixed while an unrevealed instance fact and the verified optimal decision
change. \\
Paired-interface equivalence & Direct--Mediated pair & 120 & Complete private
records reconstruct the same $\mathcal I$ while retaining different relational
structures. \\
Release integrity & World bundle & 240 & The released bundle matches the
validated inventory, manifest entries, and content digests. \\
\bottomrule
\end{tabular}%
}
\caption{Offline validation in \algoworlds. Every listed unit passes its row's
criterion. Counts use the validation unit shown in each row.}
\label{tab:validation-summary}
\endgroup
\end{table}

\section{Metric Accounting and Secondary Diagnostics}
\label{app:metrics}

The three final-decision metrics and two trajectory diagnostics are defined in
\cref{sec:metrics}. This section explains how trajectory information is
computed and how the five measures are aggregated. It then defines the
secondary diagnostics used in the supplementary analyses.

\begin{table}[H]
\centering
\begingroup
\awtablestandard
\awtablezebra{2}
\resizebox{0.90\textwidth}{!}{%
\begin{tabular}{@{}
  >{\raggedright\arraybackslash}p{3.1cm}
  >{\raggedright\arraybackslash}p{5.0cm}
  >{\raggedright\arraybackslash}p{5.4cm}@{}}
\toprule
\awtableheader
Quantity & Per-evaluation value & Within-trial summary \\
\midrule
Exact optimality & $E_w(y)\in\{0,1\}$; one only for a feasible decision at the
verified optimum. & Percentage with value one. \\
Feasibility & $F_w(y)\in\{0,1\}$; one when the family-specific checker accepts
the submitted decision. & Percentage with value one. \\
Reference utility & $U_w(y)\in[0,100]$, defined in
\cref{eq:reference-utility}. & Arithmetic mean on its $0$--$100$ scale. \\
Information sufficiency & $C_w(\tau)\in\{0,1\}$; one when the information-tool
responses contain at least one sufficient information set in full. &
Percentage with value one. \\
Discovery coverage & $D_w(\tau)\in[0,1]$; the largest fraction of any
sufficient information set contained in those responses. & Arithmetic mean
multiplied by 100. \\
\bottomrule
\end{tabular}%
}
\caption{Per-evaluation values and within-trial aggregation for the three
final-decision metrics and two trajectory diagnostics.}
\label{tab:metrics}
\endgroup
\end{table}

\paragraph{Missing final decisions.}
The three final-decision metrics in \cref{tab:metrics} follow the convention in
\cref{sec:metrics}. If no final decision is submitted, we write $y=\bot$ and
set $F_w(\bot)=E_w(\bot)=U_w(\bot)=0$. An evaluation record labeled as a
protocol violation likewise has no scorable final decision. It retains a
separate empirical outcome label but receives zero for all three metrics.
Objective value $v_w(y)$, optimality gap $\Delta_w(y)$, and normalized regret,
defined below, are available only for feasible submitted decisions. Trajectory
diagnostics remain computable from information-tool responses returned before
the evaluation ends.

\paragraph{Computing the trajectory information set.}
Let $\mathcal R_{\mathrm{fact}}(\tau)$ be the ordered sequence of executed
fact-revealing call--response records in $\tau$. Applying the response
normalizer from \cref{app:acquisition} gives the trajectory information set
used in \cref{sec:metrics}:
\begin{equation}
  \mathcal A_w(\tau)
  =\operatorname{NF}_w\!\left(\mathcal R_{\mathrm{fact}}(\tau)\right).
  \label{eq:trajectory-normalization}
\end{equation}
The normalizer combines records across responses, performs the required joins,
removes interface-specific content, and deduplicates the resulting instance
facts. The diagnostics therefore compare instance information rather than raw
text or record overlap. Candidate-evaluation calls, the final-decision tool,
and calls rejected before execution do not enter
$\mathcal R_{\mathrm{fact}}(\tau)$ and do not contribute to
$\mathcal A_w(\tau)$. If no fact-revealing call is executed,
$\mathcal A_w(\tau)=\varnothing$ and both discovery coverage and information
sufficiency are zero.

\paragraph{Aggregation.}
For a reported slice within trial $r$, let $N_r$ be its number of evaluation
records, and let $E_e,F_e,U_e,C_e$, and $D_e$ denote the five per-evaluation
values for record $e$. The trial-level summaries are
\begin{equation}
\begin{aligned}
  \overline Z_r&=\frac{100}{N_r}\sum_{e=1}^{N_r} Z_e,
    &&Z\in\{E,F,C\},\\
  \overline D_r&=\frac{100}{N_r}\sum_{e=1}^{N_r}D_e,
  &\qquad
  \overline U_r&=\frac{1}{N_r}\sum_{e=1}^{N_r}U_e.
\end{aligned}
\label{eq:headline-aggregation}
\end{equation}
Each statistic is first computed within a trial. A full trial contains 240
evaluations. Within that trial, a task-family slice contains 24 evaluations, a
workload-level slice contains 60, and a Direct or Mediated interface slice
contains 120. Thus $N_r$ is 240, 24, 60, or 120 for these summaries. The
benchmark-wide results in
\cref{tab:main-results} report the arithmetic mean and sample standard
deviation of the three trial-level values. Supplementary tables likewise use
trial-first aggregation. Each caption states the cross-trial summary shown;
not every supplementary table reports a sample standard deviation.

\paragraph{Informative trajectories and conditional outcomes.}
Following \cref{sec:main-results}, a trajectory is \emph{informative} if and
only if $C_w(\tau)=1$. The conditional analysis includes evaluations whose
trajectories satisfy this condition. Their final decisions are divided into
three groups: globally optimal when $E_w(y)=1$; feasible but suboptimal when
$F_w(y)=1$ and $E_w(y)=0$; and other failure otherwise. The last group combines
infeasible decisions, absent final decisions, and protocol violations.

Conditional exact optimality given information sufficiency is computed within
a trial as
\begin{equation}
  100\,
  \frac{\sum_{e=1}^{N_r}C_eE_e}
       {\sum_{e=1}^{N_r}C_e},
  \label{eq:conditional-exactness}
\end{equation}
when the denominator is positive. The other conditional outcome percentages
use the same within-trial denominator. Each percentage is computed within a
trial and then averaged across trials. Information sufficiency concerns the
information contained in the tool responses; it does not establish that the
agent recognized or correctly used that information. These percentages
describe final outcomes within the information-sufficient subset rather than a
causal effect of information acquisition.

\paragraph{Normalized regret.}
For a feasible decision, the secondary decision-quality diagnostic
\emph{normalized regret} is
\[
  R_w(y)=\frac{\Delta_w(y)}{\Delta_w^{\mathrm{ref}}}.
\]
It equals zero at the optimum and lies strictly between zero and one for a
non-optimal decision better than the fixed reference. It equals one for any
decision with the same gap as the reference and exceeds one for a feasible
decision with a larger gap. Within each trial, median and 90th-percentile
regret are computed among feasible non-exact decisions. Later tables report
the arithmetic mean of the three trial-specific quantiles.

For any positively scaled and translated objective
$v'_w(y)=\alpha v_w(y)+\beta$ with $\alpha>0$, both the decision's optimality
gap and $\Delta_w^{\mathrm{ref}}$ are multiplied by $\alpha$. Both normalized
regret and reference utility are therefore unchanged.

\paragraph{Entity grounding.}
As a secondary trajectory-level diagnostic, entity grounding compares the
identifiers in a submitted final decision with those previously returned to
the model. Identifier normalization maps the decision and response fields to
the canonical entity identifiers used by the world. Entity grounding is the
fraction of normalized decision identifiers that also occur in model-visible
tool responses before the final-decision call and is reported as a percentage.
It measures identifier overlap rather than whether the responses contain a
sufficient information set or whether the agent correctly uses the associated
costs and constraints.

\paragraph{Executed calls and spend.}
Let $\mathcal K_{\mathrm{exec}}(\tau)$ index the executed calls in $\tau$ whose
tools belong to $\mathcal T_{\mathrm{info}}(w)$. Executed information-tool
calls and spend are
\begin{equation}
  N_{\mathrm{call}}(\tau)=|\mathcal K_{\mathrm{exec}}(\tau)|,
  \qquad
  \operatorname{Spend}_w(\tau)
  =\sum_{i\in\mathcal K_{\mathrm{exec}}(\tau)} f_c(t_i),
  \qquad
  B_w(\tau)=\frac{\operatorname{Spend}_w(\tau)}{b}.
  \label{eq:tool-accounting}
\end{equation}
An executed candidate-evaluation call is included. The zero-cost final-decision
tool is excluded. A call rejected before execution does not increment
$N_{\mathrm{call}}$ and contributes no spend or normalized instance fact.

Within each trial, calls and spend are averaged across evaluation records.
Spend relative to budget is the mean of the per-world percentages
$100B_w(\tau)$, not a ratio formed after averaging spend and budgets.
Conditional spend is averaged over evaluations satisfying the stated
condition. The trial-level summaries are then averaged across the three
trials. These quantities describe tool use and expenditure; they are not an
efficiency score.

\paragraph{Termination and operational records.}
Two termination indicators record whether a trajectory reaches the
information-call cap or the budget cap. Each indicator is attached to the
finalized evaluation record and counted within a 240-world trial. Later tables
report the mean of the three trial-level counts. These counts are neither
percentages nor final-decision outcome categories.

\section{Executed-Work Accounting and Calibration}
\label{app:work}

\Cref{sec:workload-foundation} defines
$\operatorname{Work}(\mathcal I)$ and the four shared workload bands. This
section defines the operation counters, summarizes the parameters that govern
each family-specific exact algorithm, and reports the operation counts of the
released hidden instances.

\paragraph{Operation categories.}
Each hidden instance receives one work record from the execution of its
construction algorithm. The independent verification algorithm and later
validation checks do not contribute to this record. \Cref{tab:work-counters}
defines the five terms in \cref{eq:algorithm-work}.

\begin{table}[H]
\centering
\begingroup
\awtablecompact
\awtablezebra{2}
\resizebox{0.80\textwidth}{!}{%
\begin{tabular}{@{}
  >{\raggedright\arraybackslash}p{3.5cm}
  >{\raggedright\arraybackslash}p{9.5cm}@{}}
\toprule
\awtableheader
Counter & Executed operation represented \\
\midrule
$N_{\mathrm{state}}$ & A recurrence, graph, frontier, subset, or mask state
visited by the exact algorithm. \\
$N_{\mathrm{transition}}$ & A predecessor, successor, assignment, or other
candidate transition examined from a state. \\
$N_{\mathrm{check}}$ & A feasibility, capacity, compatibility, residue, or
other constraint condition checked during the computation. \\
$N_{\mathrm{update}}$ & An update to a stored objective or dynamic-programming
value. \\
$N_{\mathrm{reconstruct}}$ & A component of the structured optimal decision
recovered during reconstruction. \\
\bottomrule
\end{tabular}%
}
\caption{The five operation categories summed by
$\operatorname{Work}(\mathcal I)$ in \cref{eq:algorithm-work}.}
\label{tab:work-counters}
\endgroup
\end{table}

These counters measure executed operations rather than the size of the full
candidate space or a worst-case bound. In Basket Assembly,
$N_{\mathrm{state}}$ counts reached masks and $N_{\mathrm{transition}}$
counts examined ticket extensions rather than treating all $2^{3n}$ masks as
visited. In Transit Routing, $N_{\mathrm{transition}}$ counts the predecessor
candidates examined by the construction algorithm. Within a
family, every hidden instance uses the same construction algorithm and counting
rules. Across families, the five categories can occur in different proportions
but receive equal unit weight.

\paragraph{Complexity parameters.}
The bounds in \cref{app:cases} identify the parameters that enlarge each exact
algorithm's state space or transition set. \Cref{tab:work-drivers} summarizes
these parameters and guides the choice of family-specific size and structural
settings. A generated instance is assigned to a workload band using its
recorded operations, not the analytic bound.

\begin{table}[H]
\centering
\begingroup
\awtablecompact
\awtablezebra{2}
\resizebox{0.95\textwidth}{!}{%
\begin{tabular}{@{}
  >{\raggedright\arraybackslash}p{2.8cm}
  >{\raggedright\arraybackslash}p{7.0cm}
  >{\raggedright\arraybackslash}p{4.2cm}@{}}
\toprule
\awtableheader
Task family & Complexity parameters & Worst-case running time \\
\midrule
Transit Routing & Stages $d$, legs per stage $k$, lines $p$, and residue
modulus $M$ & $O(p^2\log p+dk^2M)$ \\
Basket Assembly & Items per group $n$ and examined ticket extensions
$\Delta$ & $O(2^{3n}\Delta)$ \\
Station Siting & Zones $n$, counter range $\bar H+1$ for $\bar H=n-H$, and
frontier width $\omega$ & $O(n(\bar H+1)2^\omega(\omega+1))$ \\
Authorization Planning & Residue modulus $M$ and authorization-graph vertex
and edge counts $V,E$ & $O(M(V+E))$ \\
Series Portfolio & Titles $n$, factor span $\omega$, and factors closing per
position $f_\omega$ & $O(n2^\omega(1+f_\omega))$ \\
Machine Layout & Machines $n$ and machine-subset states & $O(n2^n)$ \\
Sequential Matching & Participants $n$, used-right subsets, and final matches
& $O(n^22^n)$ \\
Fleet Dispatch & Jobs $n$, vehicles $m$, and mixed-radix load space
$P=\prod_v(C_v+1)$ & $O(nmP)$ \\
Evidence-Joined Routing & Residue modulus $M$ and graph counts
$n_V,n_E$ & $O(M(n_V+n_E))$ \\
Migration Portfolio & Support domains $n_D$ and per-mask work bound $I$ &
$O(I2^{n_D})$ \\
\bottomrule
\end{tabular}%
}
\caption{Complexity parameters and worst-case running times for the
family-specific exact algorithms. Workload levels use recorded operations
rather than these bounds.}
\label{tab:work-drivers}
\endgroup
\end{table}

\paragraph{Realized calibration.}
For each family, construction fixes one size and structural setting for each
workload level. The operation count from the construction algorithm must fall
within the corresponding band in \cref{sec:workload-foundation} and increase
strictly across levels within the family. These settings and checks are fixed
before model evaluation; model outcomes do not enter the calibration.

\begin{table}[H]
\centering
\begingroup
\awtablestandard
\awtablezebra{2}
\resizebox{0.95\textwidth}{!}{%
\begin{tabular}{@{}>{\raggedright\arraybackslash}p{3.1cm}*{4}{c}@{}}
\toprule
\awtableheader
Task family & Level 1 & Level 2 & Level 3 & Level 4 \\
\midrule
Transit Routing & 50,737--54,053 & 1,132,766--1,201,734 & 8,922,755--9,191,138 & 45,715,806--47,619,285 \\
Basket Assembly & 35,646--56,392 & 773,416--1,558,838 & 6,430,549--7,063,570 & 32,683,874--39,098,995 \\
Station Siting & 55,391--55,424 & 1,196,297--1,214,030 & 5,829,214--6,245,492 & 34,289,268--34,318,974 \\
Authorization Planning & 43,747--44,176 & 994,240--999,048 & 7,036,179--7,045,627 & 35,238,569--35,313,268 \\
Series Portfolio & 54,587 & 973,647 & 6,698,087 & 34,402,415 \\
Machine Layout & 111,668--113,178 & 1,093,813--1,107,918 & 10,405,214--10,470,928 & 46,053,123--46,226,237 \\
Sequential Matching & 107,654--108,839 & 1,446,260--1,463,310 & 7,787,593--7,824,252 & 39,493,414--39,865,720 \\
Fleet Dispatch & 42,042--47,089 & 629,346--656,846 & 4,416,131--6,235,068 & 42,862,082--53,964,927 \\
Evidence-Joined Routing & 88,935--88,999 & 1,472,138--1,472,331 & 10,616,092--10,621,525 & 52,824,682--52,827,071 \\
Migration Portfolio & 61,393--64,262 & 945,495--1,095,815 & 8,509,678--9,384,131 & 40,832,977--46,070,264 \\
\bottomrule
\end{tabular}%
}
\caption{Executed algorithmic-work ranges over the three hidden instances at
each family--workload-level setting. A singleton is shown when all three
records coincide.}
\label{tab:work-ranges}
\endgroup
\end{table}

Each cell in \cref{tab:work-ranges} gives the range over the three hidden
instances at one family--level setting. A singleton is shown when all three
records coincide. The paired worlds for one hidden instance share its work
record rather than contributing separate executions of the exact algorithm.
Across the released instances, the displayed values range from 35,646 to
53,964,927 operations, a factor of about $1.5\times10^3$.

\paragraph{Interpretive scope.}
The work record sums five equally weighted operation categories from the
family-specific construction algorithm. This record is reproducible for the
released benchmark but depends on the algorithm and accounting convention. It
does not measure wall-clock time, information-tool use, access cost, or an
algorithm-independent lower bound on the work needed to solve an instance. A
different exact algorithm or accounting convention could require a different
calibration. The relationship between workload level and LLM performance is
reported in \cref{sec:task-analysis,app:workload-results}.

\section{Operational Protocol and Record Integrity}
\label{app:operations}

An evaluation is one execution of a fixed model configuration on one
algorithmic world in one trial. Each model is evaluated in three trials over
the same 240 worlds, comprising 120 hidden instances presented through the
Direct and Mediated interfaces. This produces 720 scheduled evaluations per
model and 5,040 across the seven models. These 5,040 evaluations reuse the
same 240 worlds across models and trials. This section records the request
settings, interaction controls, and evaluation-record accounting used in the
experiments. Integrity of the benchmark world bundles is described separately
in \cref{app:release-integrity}.

The trajectory in \cref{eq:trajectory} gives the interaction form when the
agent submits a final decision. A finalized trajectory may instead end without
invoking $t_{\mathrm{dec}}$; the metric convention for that case is given in
\cref{app:metrics}.

\begin{table}[H]
\centering
\begingroup
\awtablestandard
\setlength{\tabcolsep}{2.25pt}
\awtablezebra{2}
\resizebox{0.95\textwidth}{!}{%
\begin{tabular}{@{}>{\raggedright\arraybackslash}p{2.35cm}>{\raggedright\arraybackslash}p{4.1cm}>{\raggedright\arraybackslash}p{5.0cm}cc@{}}
\toprule
\awtableheader
Model & Model ID / API & Thinking / reasoning & Output cap & Temperature \\
\midrule
GPT-5.6 Terra & \nolinkurl{gpt-5.6-terra}\newline \textit{Responses API} & reasoning.effort=max & 128K & Omitted \\
GPT-5.6 Sol & \nolinkurl{gpt-5.6-sol}\newline \textit{Responses API} & reasoning.effort=max & 128K & Omitted \\
Qwen 3.5 Plus & \nolinkurl{qwen3.5-plus}\newline \textit{Chat Completions API} & thinking=true; budget=81,920 & 65,536 & 0.6 \\
DeepSeek V4 Pro & \nolinkurl{deepseek-v4-pro-202606}\newline \textit{Chat Completions API} & thinking=true; reasoning\_effort=max & 128K & Omitted \\
Claude Opus 4.8 & \nolinkurl{claude-opus-4-8}\newline \textit{Anthropic Messages API} & adaptive thinking; effort=max & 128K & Omitted \\
Claude Sonnet 5 & \nolinkurl{claude-sonnet-5}\newline \textit{Anthropic Messages API} & adaptive thinking; effort=max & 128K & Omitted \\
GLM 5.2 & \nolinkurl{glm-5.2}\newline \textit{Chat Completions API} & thinking=true; reasoning\_effort=max & 128K & 1.0 \\
\bottomrule
\end{tabular}%
}
\caption{Model-specific request settings used in the reported experiments. The
output cap is the returned-output token limit applied to each model request,
not a trajectory-level token budget, and is distinct from any internal thinking
budget. Here, K denotes 1,024 tokens, and Qwen's listed thinking budget is also
measured in tokens. All models use a configuration with thinking or reasoning
enabled. Temperature is marked \emph{Omitted} when no explicit temperature
value was supplied in the reported request configuration.}
\label{tab:subjects}
\endgroup
\end{table}

\paragraph{Fixed request configurations.}
All seven models use fixed configurations with the provider's reasoning
feature enabled; some providers call this feature thinking. Before the full
evaluation, we verify each configuration with a protocol-compatibility run
that uses tools over multiple rounds and makes a valid structured final
submission. This run confirms that the configuration can follow the
interaction protocol; it does not require every subsequent evaluation to end
with a valid decision.

\Cref{tab:subjects} reports the exact model identifiers, API families,
reasoning or thinking controls, output-token limits, and temperatures used
in the experiments. Provider-specific thinking and reasoning controls are
reported as configured. They are not treated as numerically equivalent compute
budgets.

\paragraph{Shared interaction controls.}
In each evaluation, the agent receives the textual question $q$, the
task-specific tool schemas $\mathcal T$, their costs under $f_c$, and the
access budget $b$. It does not receive the hidden instance $\mathcal I$ or
offline benchmark fields such as the verified optimum and sufficient
information sets. Parallel tool calls are disabled, so each response is
recorded before the next call is issued.

Each trajectory permits at most 255 executed information-tool calls in
addition to its world-specific access budget. The information-call cap uses
$N_{\mathrm{call}}(\tau)$ from \cref{eq:tool-accounting}. An executed
candidate-evaluation call counts toward this cap; the final-decision tool and
calls rejected before execution do not. A rejected call also incurs no
benchmark cost. The zero-cost final-decision tool may be invoked at most once
and ends the trajectory.

The agent-facing tool suite contains neither a general-purpose code execution
environment nor an optimization solver. In particular, the exact algorithms
used for benchmark construction and validation are not exposed to the
evaluated agents.

\paragraph{Scheduled evaluations, attempts, and finalization.}
Each model configuration--world--trial combination defines one scheduled
evaluation. The reported results contain exactly one finalized record for
every scheduled evaluation. This record includes the trajectory and its
metrics. If it contains no final-decision submission, we write $y=\bot$ and
apply the metric convention in \cref{sec:metrics}.

Provider and network attempts are individual request records linked to the
scheduled evaluation. They are not trajectory outcomes and do not enter the
final-decision outcome or cap-termination counts.
Information-call-cap and budget-cap terminations are status indicators on the
finalized record. Absent final decisions and protocol violations are instead
final-decision outcome categories.

\paragraph{Manifest integrity.}
The benchmark manifest identifies the validated 240-world inventory through
manifest entries and content digests, as described in
\cref{app:release-integrity}. The
experiment manifest associates each finalized record with its model, trial,
world, and protocol and links any provider or network attempts. Before table
generation, integrity checks verify these identifiers and content digests and
require 240 finalized records for every model--trial pair. The empirical tables
use only records that match these manifest entries and content digests.

\section{Supplementary Experimental Results}
\label{app:full-results}

This section provides detailed outcome, workload, interface, and trajectory
analyses that support the results in \cref{sec:experiments}.

\begin{figure}[H]
\centering
\begin{tikzpicture}
\begin{axis}[
  width=\linewidth,height=5.8cm,
  xmin=-48,xmax=105,ymin=0.45,ymax=7.75,
  xlabel={Mean rate (\%)},
  xtick={0,20,40,60,80,100},
  ytick={1,2,3,4,5,6,7},
  yticklabels={{GPT-5.6 Terra},{GPT-5.6 Sol},{Qwen 3.5 Plus},{DeepSeek V4 Pro},{Claude Opus 4.8},{Claude Sonnet 5},{GLM 5.2}},
  axis y line*=left,axis x line*=bottom,
  y axis line style={draw=none},ytick style={draw=none},
  xmajorgrids=true,grid style={black!8},
  tick label style={font=\scriptsize},
  yticklabel style={anchor=west,xshift=2pt,font=\tiny},
  label style={font=\scriptsize},
  legend columns=2,
  legend style={font=\scriptsize,draw=none,at={(0.5,1.03)},anchor=south},
]
\addplot[black!22,line width=1.2pt] coordinates {(26.3889,1) (79.7222,1)};
\addplot[black!22,line width=1.2pt] coordinates {(38.1944,2) (89.8611,2)};
\addplot[black!22,line width=1.2pt] coordinates {(5.4167,3) (37.9167,3)};
\addplot[black!22,line width=1.2pt] coordinates {(16.1111,4) (79.5833,4)};
\addplot[black!22,line width=1.2pt] coordinates {(38.6111,5) (81.5278,5)};
\addplot[black!22,line width=1.2pt] coordinates {(32.3611,6) (86.3889,6)};
\addplot[black!22,line width=1.2pt] coordinates {(24.8611,7) (83.3333,7)};
\addplot[only marks,mark=*,mark size=2.5pt,AWgold,point meta=explicit symbolic,nodes near coords,nodes near coords align={horizontal},every node near coord/.append style={font=\tiny,anchor=south west,text=AWgold,xshift=1pt}] coordinates {(26.3889,1) [26.4] (38.1944,2) [38.2] (5.4167,3) [5.4] (16.1111,4) [16.1] (38.6111,5) [38.6] (32.3611,6) [32.4] (24.8611,7) [24.9]};
\addlegendentry{Exact optimality}
\addplot[only marks,mark=*,mark size=2.5pt,AWteal,point meta=explicit symbolic,nodes near coords,nodes near coords align={horizontal},every node near coord/.append style={font=\tiny,anchor=south east,text=AWteal,xshift=-1pt}] coordinates {(79.7222,1) [79.7] (89.8611,2) [89.9] (37.9167,3) [37.9] (79.5833,4) [79.6] (81.5278,5) [81.5] (86.3889,6) [86.4] (83.3333,7) [83.3]};
\addlegendentry{Information sufficiency}
\end{axis}
\end{tikzpicture}
\caption{Mean exact-optimality and information-sufficiency rates over three
trials on the same fixed worlds. Each line connects the two overall rates for
one model. The figure does not condition final decisions on information
sufficiency within the same evaluation. Across-trial standard deviations
appear in \cref{tab:main-results}, and repeated-trial profiles appear in
\cref{tab:stability}.}
\label{fig:results-overview}
\end{figure}
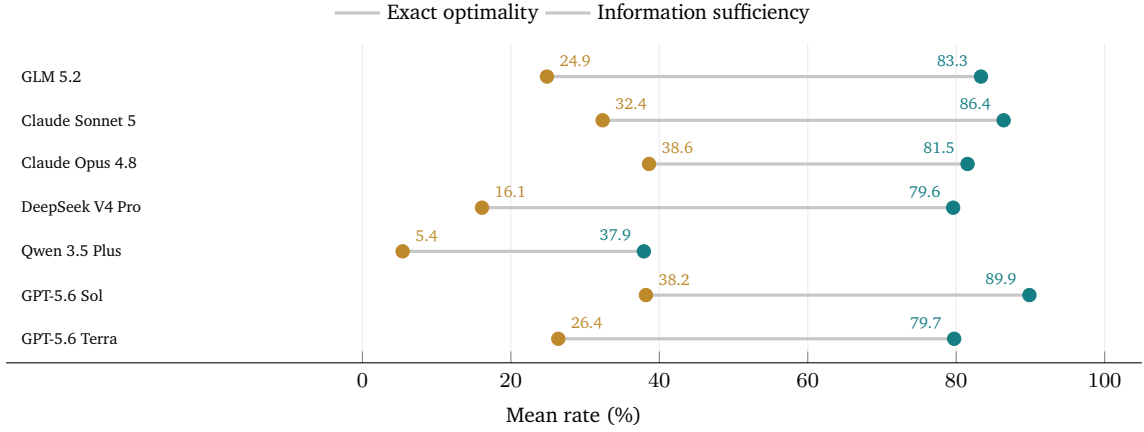

The corresponding within-evaluation analysis appears in
\cref{sec:main-results,tab:conditional-outcomes}.

\subsection{Final-Decision Outcomes and Quality}

\begin{table}[H]
\centering
\begingroup
\awtablestandard
\setlength{\tabcolsep}{7pt}
\awtablezebra{2}
\resizebox{0.80\textwidth}{!}{%
\begin{tabular}{@{}lccccc@{}}
\toprule
\awtableheader
\textbf{Model} & \makecell{\textbf{Globally}\\\textbf{optimal}} & \makecell{\textbf{Feasible}\\\textbf{suboptimal}} & \textbf{Infeasible} & \makecell{\textbf{No}\\\textbf{decision}} & \makecell{\textbf{Protocol}\\\textbf{violation}} \\
\midrule
GPT-5.6 Terra & 63.3 & 147.3 & 29.3 & 0.0 & 0.0 \\
GPT-5.6 Sol & 91.7 & 142.3 & 6.0 & 0.0 & 0.0 \\
Qwen 3.5 Plus & 13.0 & 104.3 & 121.7 & 0.0 & 1.0 \\
DeepSeek V4 Pro & 38.7 & 162.7 & 33.7 & 5.0 & 0.0 \\
Claude Opus 4.8 & 92.7 & 138.3 & 5.7 & 3.3 & 0.0 \\
Claude Sonnet 5 & 77.7 & 143.3 & 17.7 & 1.3 & 0.0 \\
GLM 5.2 & 59.7 & 95.3 & 15.0 & 58.7 & 11.3 \\
\bottomrule
\end{tabular}%
}
\caption{Mean counts by final-decision category per 240-world trial, averaged
across the three trials. The five categories partition the evaluation records
before rounding. Protocol violations remain separate from infeasible structured
decisions and absent final decisions.}
\label{tab:outcomes}
\endgroup
\end{table}

\begin{table}[H]
\centering
\begingroup
\awtablestandard
\setlength{\tabcolsep}{2.0pt}
\awtablezebra{2}
\resizebox{0.98\textwidth}{!}{%
\begin{tabular}{@{}>{\raggedright\arraybackslash}p{2.7cm}*{7}{c}@{}}
\toprule
\awtableheader
\textbf{Model} & \makecell{\textbf{Reference}\\\textbf{utility}} & \makecell{\textbf{Median}\\\textbf{regret}} & \makecell{\textbf{90th}\\\textbf{percentile}\\\textbf{regret}} & \makecell{\textbf{Globally}\\\textbf{optimal}} & \makecell{\textbf{Better than}\\\textbf{reference}} & \makecell{\textbf{Reference}\\\textbf{or worse}} & \makecell{\textbf{Infeasible,}\\\textbf{absent, or}\\\textbf{protocol}\\\textbf{violation}} \\
\midrule
GPT-5.6 Terra & 54.10 $\pm$ 1.95 & 0.526 & 1.582 & 63.3 & 91.3 & 56.0 & 29.3 \\
GPT-5.6 Sol & 74.98 $\pm$ 1.77 & 0.214 & 1.000 & 91.7 & 119.3 & 23.0 & 6.0 \\
Qwen 3.5 Plus & 15.82 $\pm$ 0.87 & 1.000 & 4.277 & 13.0 & 30.7 & 73.7 & 122.7 \\
DeepSeek V4 Pro & 28.49 $\pm$ 0.91 & 1.000 & 2.582 & 38.7 & 55.7 & 107.0 & 38.7 \\
Claude Opus 4.8 & 73.14 $\pm$ 11.81 & 0.354 & 1.000 & 92.7 & 107.0 & 31.3 & 9.0 \\
Claude Sonnet 5 & 67.20 $\pm$ 9.53 & 0.357 & 0.973 & 77.7 & 118.7 & 24.7 & 19.0 \\
GLM 5.2 & 35.48 $\pm$ 2.67 & 1.000 & 1.655 & 59.7 & 41.0 & 54.3 & 85.0 \\
\bottomrule
\end{tabular}%
}
\caption{Reference-normalized decision quality. Utility is the mean $\pm$
sample standard deviation across three trial-level means; regret entries
average the trial-specific quantiles among feasible non-exact decisions; and
outcome columns are mean counts per 240-world trial. Better than reference
means normalized regret strictly between zero and one, whereas reference or
worse means normalized regret at least one. The four outcome columns partition
each trial before rounding; the final column combines infeasible, absent, and
protocol-violation outcomes.}
\label{tab:reference-quality}
\endgroup
\end{table}

\paragraph{Outcome categories.}
\Cref{tab:outcomes} partitions all evaluation records by final-decision
outcome. Feasible but suboptimal decisions are the largest category for six of
the seven models. Qwen is the exception, with more infeasible than feasible but
suboptimal decisions. GLM has substantially more absent final decisions than
the other models.

\paragraph{Quality among feasible suboptimal decisions.}
\Cref{tab:reference-quality} reports reference utility and outcome counts over
all evaluations. Its normalized-regret quantiles are instead computed among
feasible non-exact decisions. The mean of the three trial-specific median
regrets is below 1.0 for Sol (0.214), Opus (0.354), Sonnet (0.357), and Terra
(0.526), and equals 1.0 for Qwen, DeepSeek, and GLM. A regret of 1.0 corresponds
to the fixed reference gap. The mean trial-specific 90th percentile is at least
1.0 for six models; Sonnet's is 0.973, whereas Qwen's is 4.277. Because the
table averages trial-specific quantiles, an entry of 1.0 does not mean that
every feasible non-exact decision has the reference gap.

\paragraph{Informative trajectories.}
\begin{table}[H]
\centering
\begingroup
\awtablestandard
\setlength{\tabcolsep}{2.25pt}
\awtablezebra{3}
\resizebox{0.90\textwidth}{!}{%
\begin{tabular}{@{}>{\raggedright\arraybackslash}p{2.8cm}*{6}{c}@{}}
\toprule
\awtableheader
& \multicolumn{1}{c}{\textbf{Conditioning count}} & \multicolumn{5}{c}{\textbf{Final-decision outcomes}} \\
\cmidrule(lr){2-2}\cmidrule(lr){3-7}
\rowcolor{white}
\textbf{Model} & \makecell{\textbf{Information}\\\textbf{sufficiency}} & \makecell{\textbf{Globally}\\\textbf{optimal}} & \makecell{\textbf{Feasible}\\\textbf{suboptimal}} & \textbf{Infeasible} & \makecell{\textbf{No}\\\textbf{decision}} & \makecell{\textbf{Protocol}\\\textbf{violation}} \\
\midrule
GPT-5.6 Terra & 191.3 & 56.0 & 114.3 & 21.0 & 0.0 & 0.0 \\
GPT-5.6 Sol & 215.7 & 89.7 & 121.3 & 4.7 & 0.0 & 0.0 \\
Qwen 3.5 Plus & 91.0 & 3.0 & 45.3 & 42.3 & 0.0 & 0.3 \\
DeepSeek V4 Pro & 191.0 & 34.0 & 132.7 & 21.3 & 3.0 & 0.0 \\
Claude Opus 4.8 & 195.7 & 71.7 & 116.3 & 4.3 & 3.3 & 0.0 \\
Claude Sonnet 5 & 207.3 & 70.7 & 120.0 & 15.3 & 1.3 & 0.0 \\
GLM 5.2 & 200.0 & 52.0 & 81.0 & 13.0 & 54.0 & 0.0 \\
\bottomrule
\end{tabular}%
}
\caption{Mean counts by final-decision category among evaluations satisfying
information sufficiency. Each count is computed within a 240-world trial and then averaged
across three trials, so the five decision columns sum to the information-sufficiency
count in each row up to rounding.}
\label{tab:conditional-outcome-counts}
\endgroup
\end{table}

As defined in \cref{app:metrics}, an informative trajectory satisfies
$C_w(\tau)=1$. \Cref{tab:conditional-outcome-counts} partitions the
corresponding evaluations into globally optimal, feasible but suboptimal,
infeasible, absent-final-decision, and protocol-violation outcomes. The
``Other failure'' column in \cref{tab:conditional-outcomes} combines the final
three categories.

Within the informative subset, feasible but suboptimal decisions are the
largest final-decision category for all seven models. Qwen averages 42.3
infeasible decisions per trial in this subset, whereas GLM averages 54.0 absent
final decisions. The detailed partition shows that these models reach the
``Other failure'' category in different ways.

\subsection{Task-Level Reference Utility and Discovery Coverage}

\begin{table}[H]
\centering
\begingroup
\awtablestandard
\setlength{\tabcolsep}{2.0pt}
\awtablezebra{3}
\resizebox{0.95\textwidth}{!}{%
\begin{tabular}{@{}>{\raggedright\arraybackslash}p{3.1cm}*{7}{c}@{}}
\toprule
\awtableheader
& \multicolumn{7}{c}{\textbf{Reference utility / discovery coverage}} \\
\cmidrule(lr){2-8}
\rowcolor{white}
Task family & Terra & Sol & Qwen & DeepSeek & Opus & Sonnet & GLM \\
\midrule
Transit Routing & 40.0/90.6 & 77.1/98.4 & 0.5/55.2 & 3.6/98.2 & 53.5/98.2 & 19.8/99.7 & 1.3/98.6 \\
Basket Assembly & 100.0/99.3 & 100.0/99.5 & 59.0/80.9 & 91.6/98.7 & 100.0/89.6 & 100.0/95.7 & 98.5/94.8 \\
Station Siting & 88.8/94.9 & 99.9/100.0 & 2.5/48.7 & 10.0/99.1 & 98.8/95.6 & 92.1/96.3 & 59.5/97.9 \\
Authorization Planning & 18.9/96.9 & 75.6/100.0 & 0.0/99.2 & 4.2/99.2 & 65.0/100.0 & 65.7/100.0 & 19.4/98.6 \\
Series Portfolio & 20.8/96.9 & 34.5/100.0 & 0.2/68.2 & 1.2/100.0 & 48.9/100.0 & 49.6/98.3 & 7.4/98.6 \\
Machine Layout & 33.1/94.1 & 56.1/95.4 & 0.0/90.4 & 0.6/92.9 & 40.0/99.7 & 38.3/96.3 & 7.1/95.8 \\
Sequential Matching & 92.7/99.6 & 95.7/99.8 & 15.9/88.0 & 66.0/90.4 & 95.6/100.0 & 98.2/100.0 & 47.0/60.4 \\
Fleet Dispatch & 99.3/100.0 & 98.6/100.0 & 80.1/99.0 & 86.9/99.6 & 99.9/100.0 & 84.2/99.7 & 91.5/95.8 \\
Evidence-Joined Routing & 24.9/94.1 & 45.2/97.9 & 0.0/86.6 & 15.4/99.9 & 43.2/90.3 & 53.4/99.9 & 1.2/98.6 \\
Migration Portfolio & 22.4/100.0 & 67.2/100.0 & 0.0/99.2 & 5.4/99.5 & 86.3/100.0 & 70.6/100.0 & 21.9/100.0 \\
\bottomrule
\end{tabular}%
}
\caption{Task-level reference utility and discovery coverage. Within each
trial, each task-family slice contains 24 worlds. Each cell reports the mean
across the three trial-level reference-utility means, followed by the mean
across the three trial-level discovery-coverage percentages. Reference utility
is on its $[0,100]$ scale, whereas discovery coverage is a percentage. These
are marginal task-level summaries of decision quality and the information
contained in the recorded tool responses.}
\label{tab:task-diagnostics}
\endgroup
\end{table}

\Cref{tab:task-diagnostics} shows that high discovery coverage can accompany
very different levels of decision quality. In Authorization Planning, Qwen and
DeepSeek both have 99.2\% discovery coverage but reference utilities of 0.0 and
4.2, respectively. In Series Portfolio, DeepSeek has 100.0\% discovery
coverage and reference utility 1.2. Fleet Dispatch provides a contrasting
profile: all seven models have discovery coverage of at least 95.8\% and
reference utility of at least 80.1.

These are marginal task-level summaries. The within-evaluation analysis based
on information sufficiency appears in
\cref{sec:main-results,tab:conditional-outcomes}. The summaries do not identify
which family properties cause the observed differences. The complexity bounds
in \cref{app:cases} describe the offline exact algorithms, not the reasoning
process of an evaluated agent.

\subsection{Workload-Level Patterns and Scope}
\label{app:workload-results}

\Cref{tab:workload-results} reports exact optimality and reference utility by
workload level. Each level contains 60 worlds within a trial: 30 hidden
instances---three from each of ten families---presented through both paired
interfaces. The three trials repeat these same fixed worlds rather than adding
new hidden instances.

The main trend is consistent across models. All seven have lower exact
optimality and reference utility at L4 than at L1. Reference utility decreases
at every successive level for six models; DeepSeek V4 Pro is the exception,
with a partial recovery from 22.3 at L3 to 28.1 at L4 while remaining below its
L1 value of 35.9. Exact-optimality rates fluctuate more across intermediate
levels, even though every model ends lower at L4 than at L1.

The levels are assigned from the work executed by the construction algorithms
before model evaluation; model outcomes do not enter their calibration
(\cref{app:work}). Every model performs worse at L4 than at L1, indicating that
this scale captures a dimension of challenge for the evaluated agents. It does not imply
strict monotonicity for every model or adjacent level. The four levels contain
different fixed instances rather than four versions of the same instance, so
adjacent-level differences may reflect instance composition. With three trials
per model, small reversals may also reflect run-to-run variation. The observed
relationship is descriptive rather than a causal effect of workload.

\subsection{Detailed Paired-Interface Accounting}
\label{app:paired-interface-results}

This subsection gives the pairing and interval construction used for
\cref{tab:representation-results}.

For a fixed model, hidden instance $i$, and trial $r$, let
$E^{(\mathrm D)}_{i,r}$ and $E^{(\mathrm M)}_{i,r}$ be the binary
exact-optimality outcomes under the Direct and Mediated interfaces, and define

\begin{equation}
  d_{i,r}=E^{(\mathrm D)}_{i,r}-E^{(\mathrm M)}_{i,r}.
  \label{eq:paired-interface-outcome}
\end{equation}

A Direct win has $d_{i,r}=1$, a Mediated win has $d_{i,r}=-1$, and a tie has
$d_{i,r}=0$. Each model contributes $120\times3=360$ matched
instance--trial pairs. A tie means only that the two evaluations share the same
binary exact-optimality status: both may be exact or both non-exact. It does not
imply identical submitted decisions, objective values, or tool trajectories.
If $W_{\mathrm D}$ and $W_{\mathrm M}$ denote the Direct- and Mediated-win
counts, respectively, the displayed rate contrast satisfies

\begin{equation}
  \Delta_{\mathrm{D-M}}
  =\frac{100}{360}\sum_{i=1}^{120}\sum_{r=1}^{3}d_{i,r}
  =100\frac{W_{\mathrm D}-W_{\mathrm M}}{360}.
  \label{eq:paired-interface-difference}
\end{equation}

For interval construction, the three paired outcomes for an instance are first
averaged as $\bar d_i=\frac{1}{3}\sum_{r=1}^{3}d_{i,r}$. The paired bootstrap
resamples the 120 values $\{\bar d_i\}_{i=1}^{120}$ with replacement 10,000
times and recomputes their percentage-point mean. Thus the resampling unit is
the hidden instance after trial averaging, not one of the 360 repeated
instance--trial pairs. The displayed contrasts are computed from unrounded
rates.

Across the 360 matched pairs for each model, Direct wins range from 11 to 34,
Mediated wins from 6 to 13, and ties from 315 to 343. A positive aggregate
contrast therefore does not imply pointwise Direct dominance. Only GPT-5.6 Sol
and Claude Sonnet 5 have 95\% intervals that exclude zero. The comparison is
limited to the 120 validated Direct--Mediated pairs in this benchmark and does
not establish a Direct advantage for other interface designs. Structural
equivalence checks for these pairs appear in \cref{app:paired-equivalence}.

\subsection{Repeated-Trial Profiles}

\begin{table}[H]
\centering
\begingroup
\awtablestandard
\setlength{\tabcolsep}{2.0pt}
\awtablezebra{3}
\resizebox{0.90\textwidth}{!}{%
\begin{tabular}{@{}>{\raggedright\arraybackslash}p{2.8cm}*{8}{c}@{}}
\toprule
\awtableheader
& \multicolumn{3}{c}{\textbf{Trial exact-optimality rate}} & \multicolumn{4}{c}{\textbf{Trials attaining exact optimality per world}} & \makecell{\textbf{Pairwise}\\\textbf{agreement}} \\
\cmidrule(lr){2-4}\cmidrule(lr){5-8}
\rowcolor{white}
Model & T1 & T2 & T3 & 0/3 & 1/3 & 2/3 & 3/3 & \\
\midrule
GPT-5.6 Terra & 27.9 & 25.4 & 25.8 & 164 & 8 & 22 & 46 & 91.7 \\
GPT-5.6 Sol & 39.2 & 38.8 & 36.7 & 135 & 16 & 8 & 81 & 93.3 \\
Qwen 3.5 Plus & 5.8 & 5.4 & 5.0 & 218 & 9 & 9 & 4 & 95.0 \\
DeepSeek V4 Pro & 17.1 & 16.2 & 15.0 & 187 & 14 & 15 & 24 & 91.9 \\
Claude Opus 4.8 & 33.3 & 40.4 & 42.1 & 127 & 18 & 25 & 70 & 88.1 \\
Claude Sonnet 5 & 27.9 & 34.6 & 34.6 & 141 & 22 & 20 & 57 & 88.3 \\
GLM 5.2 & 24.2 & 25.0 & 25.4 & 166 & 8 & 27 & 39 & 90.3 \\
\bottomrule
\end{tabular}%
}
\caption{Repeated-execution profiles on the same fixed worlds in \algoworlds.
T1--T3 are exact-optimality rates in percent. Columns 0/3 through 3/3 count
worlds on which the model attains exact optimality in that many trials.
For each pair of trials, pairwise agreement is the percentage of the 240 fixed
worlds whose binary exact-optimality statuses agree; the table reports the mean
over the three trial pairs. The four frequency counts sum to 240 worlds, and
together with pairwise agreement describe binary exact-optimality repeatability
on this fixed suite, not agreement in objective value or submitted decision.}
\label{tab:stability}
\endgroup
\end{table}

Two summaries describe variation across the three trials. First, the
trial-level exact-optimality rates in \cref{tab:main-results} have sample
standard deviations of 0.42--1.34 percentage points for GPT-5.6 Terra,
GPT-5.6 Sol, Qwen 3.5 Plus, DeepSeek V4 Pro, and GLM 5.2. The corresponding
values are 4.65 and 3.85 points for Claude Opus 4.8 and Claude Sonnet 5.
Second, \cref{tab:stability} holds the model and algorithmic world fixed and
compares binary exact-optimality status across trials. Mean pairwise agreement
ranges from 88.1\% to 95.0\% across models.

Qwen has the highest pairwise agreement, at 95.0\%, yet it never attains exact
optimality on 218 of the 240 worlds and does so in all three trials on only
four. GPT-5.6 Sol has 93.3\% pairwise agreement and attains exact optimality in
all three trials on 81 worlds. Pairwise agreement therefore measures
repeatability of the binary exact-optimality status; it is neither an
exact-optimality rate nor a measure of agreement in submitted decisions or
objective values.

\subsection{Additional Tool-Use and Termination Diagnostics}

\begin{table}[H]
\centering
\begingroup
\awtablestandard
\setlength{\tabcolsep}{6.5pt}
\awtablezebra{3}
\resizebox{1.00\textwidth}{!}{%
\begin{tabular}{@{}lccccccc@{}}
\toprule
\awtableheader
& \multicolumn{3}{c}{\textbf{Trajectory-level summaries}} & \multicolumn{1}{c}{\textbf{Tool use}} & \multicolumn{3}{c}{\makecell{\textbf{Conditional}\\\textbf{and termination}\\\textbf{summaries}}} \\
\cmidrule(lr){2-4}\cmidrule(lr){5-5}\cmidrule(lr){6-8}
\rowcolor{white}
Model & Discovery & Grounding & Sufficiency & Calls & \makecell{Exact given\\sufficiency} & \makecell{Info.-call\\cap} & \makecell{Budget\\cap} \\
\midrule
GPT-5.6 Terra & 96.65 & 100.00 & 79.72 & 54.9 & 29.24 & 0.0 & 0.0 \\
GPT-5.6 Sol & 99.10 & 100.00 & 89.86 & 57.4 & 41.57 & 0.0 & 0.0 \\
Qwen 3.5 Plus & 81.55 & 97.91 & 37.92 & 34.6 & 3.31 & 0.0 & 0.0 \\
DeepSeek V4 Pro & 97.74 & 97.27 & 79.58 & 69.0 & 17.82 & 2.0 & 0.7 \\
Claude Opus 4.8 & 97.33 & 98.60 & 81.53 & 58.2 & 36.33 & 3.3 & 0.0 \\
Claude Sonnet 5 & 98.59 & 99.44 & 86.39 & 59.9 & 33.92 & 1.3 & 0.0 \\
GLM 5.2 & 93.91 & 70.82 & 83.33 & 56.8 & 25.97 & 57.3 & 1.3 \\
\bottomrule
\end{tabular}%
}
\caption{Trajectory-level, tool-use, conditional, and termination summaries.
Discovery coverage and information sufficiency are the two trajectory
diagnostics defined in \cref{sec:metrics}; entity grounding is a secondary
diagnostic defined in \cref{app:metrics}. Each numerical entry is the mean of
the three trial-level summaries. These three quantities and conditional exact
optimality are percentages. Calls are mean executed information-tool calls per
evaluation; information-call- and budget-cap terminations are mean counts per
240-world trial.}
\label{tab:diagnostics}
\endgroup
\end{table}

\paragraph{Entity grounding.}
Six models have entity grounding between 97.27\% and 100.00\%. GLM differs
from this pattern: its mean discovery coverage is 93.91\%, while its mean
entity grounding is 70.82\%. Discovery coverage measures overlap with
sufficient information sets, whereas entity grounding measures identifier
overlap with earlier tool responses. Neither quantity shows whether the agent
correctly used the returned costs and constraints.

\paragraph{Cap terminations.}
GLM reaches the information-call cap in a mean of 57.3 evaluations per
240-world trial. Mean information-call-cap counts are also nonzero for
DeepSeek, Claude Opus 4.8, and Claude Sonnet 5, at 2.0, 3.3, and 1.3,
respectively. Mean budget-cap counts are nonzero only for DeepSeek and GLM, at
0.7 and 1.3 per trial. These termination indicators are separate from
final-decision categories and from the provider and network attempts described
in \cref{app:operations}.

\begin{table}[H]
\centering
\begingroup
\awtablestandard
\setlength{\tabcolsep}{12pt}
\awtablezebra{3}
\resizebox{0.85\textwidth}{!}{%
\begin{tabular}{@{}lccccc@{}}
\toprule
\awtableheader
& \multicolumn{3}{c}{\textbf{Overall acquisition}} & \multicolumn{2}{c}{\textbf{Conditional spend}} \\
\cmidrule(lr){2-4}\cmidrule(lr){5-6}
\rowcolor{white}
\textbf{Model} & \textbf{Calls} & \textbf{Spend} & \makecell{\textbf{Spend}\\\textbf{/ budget}} & \makecell{\textbf{Given}\\\textbf{information}\\\textbf{sufficiency}} & \makecell{\textbf{Given}\\\textbf{exact}\\\textbf{optimality}} \\
\midrule
GPT-5.6 Terra & 54.9 & 335.4 & 68.1\% & 360.1 & 367.3 \\
GPT-5.6 Sol & 57.4 & 448.2 & 80.6\% & 464.0 & 323.8 \\
Qwen 3.5 Plus & 34.6 & 291.9 & 66.5\% & 273.4 & 311.7 \\
DeepSeek V4 Pro & 69.0 & 424.5 & 83.0\% & 455.1 & 335.4 \\
Claude Opus 4.8 & 58.2 & 367.6 & 73.3\% & 353.7 & 328.9 \\
Claude Sonnet 5 & 59.9 & 369.8 & 74.8\% & 373.3 & 357.7 \\
GLM 5.2 & 56.8 & 357.2 & 67.6\% & 390.4 & 325.4 \\
\bottomrule
\end{tabular}%
}
\caption{Descriptive acquisition accounting. Calls and spend follow
\cref{eq:tool-accounting}: they include candidate-evaluation calls when executed and
exclude the final-decision tool and calls rejected before execution. Calls,
spend, and spend relative to budget are computed per evaluation, averaged
within each trial, and then averaged across three trials. Spend and both
conditional-spend columns are in benchmark cost units; spend/budget is instead
the mean of the per-world percentages. Conditional columns are likewise
computed within each trial for evaluations satisfying information sufficiency
or exact optimality and then averaged across the three trials; they are not
efficiency scores or estimates of a spending effect.}
\label{tab:acquisition-spend}
\endgroup
\end{table}

\paragraph{Executed calls and spend.}
\Cref{tab:acquisition-spend} reports executed calls and expenditure in
benchmark cost units. It keeps absolute spend separate from the mean per-world
spend-to-budget percentage. Qwen has the lowest mean call count, absolute
spend, and budget percentage, at 34.6, 291.9, and 66.5\%, respectively. Sol has
the highest absolute spend at 448.2, whereas DeepSeek has the highest mean
per-world budget percentage at 83.0\%.

Conditional spend follows different patterns across models. For Sol, the
overall mean, the mean among information-sufficient evaluations, and the mean
among exact-optimal evaluations are 448.2, 464.0, and 323.8, respectively. For
Qwen, the corresponding values are 291.9, 273.4, and 311.7. These conditional
summaries describe the selected subsets; they are not efficiency scores or
estimates of a spending effect.

\subsection{Structured Trajectory Examples}
\label{app:trajectories}

\begin{table}[H]
\centering
\begingroup
\awtablestandard
\awtablezebra{2}
\resizebox{0.95\textwidth}{!}{%
\begin{tabular}{@{}>{\raggedright\arraybackslash}p{2.5cm}>{\raggedright\arraybackslash}p{13.0cm}@{}}
\toprule
\awtableheader
Field & Recorded trajectory \\
\midrule
\rowcolor{AWblue!7}
\textbf{Archetype} & \textbf{Tool responses do not contain a sufficient information set in full} \\
World & Station Siting, Mediated, L2, instance 1; DeepSeek V4 Pro, trial 2 \\
Recorded calls & read zone window (8); read zone (3); read demand (11) \\
Final result & No final decision; discovery coverage 35.0\%. \\
\addlinespace[2pt]\midrule\addlinespace[2pt]
\rowcolor{AWblue!7}
\textbf{Archetype} & \textbf{Sufficient information, infeasible} \\
World & Evidence-Joined Routing, Mediated, L1, instance 2; GPT-5.6 Terra, trial 1 \\
Recorded calls & read tariff card (1); read manifest window (4); read exceptions window (4); read manifest stage (1); read exceptions stage (1); submit path (1) \\
Final result & Final decision infeasible; discovery coverage and entity grounding both 100.0\%. \\
\addlinespace[2pt]\midrule\addlinespace[2pt]
\rowcolor{AWblue!7}
\textbf{Archetype} & \textbf{Sufficient information, feasible suboptimal} \\
World & Series Portfolio, Direct, L2, instance 3; Claude Sonnet 5, trial 1 \\
Recorded calls & read catalog window (22); read series page (20); submit portfolio (1) \\
Final result & Feasible non-optimum; normalized regret 0.507, reference utility 49.3. \\
\bottomrule
\end{tabular}%
}
\caption{Three examples drawn from recorded trajectories. The recorded-call
rows give a compact inventory of tool names and counts in first-use order,
include the final-decision submission when one occurs, and exclude private
reasoning text. They do not reproduce the full call sequence and are not the
$N_{\mathrm{call}}$ diagnostic, which excludes $t_{\mathrm{dec}}$. The examples
illustrate distinct information-availability and final-decision outcomes;
corresponding aggregate results appear in
\cref{tab:main-results,tab:conditional-outcomes,tab:reference-quality,tab:diagnostics}.
They are illustrative examples rather than estimates of outcome prevalence.}
\label{tab:trajectory-examples}
\endgroup
\end{table}

The three examples illustrate the gap between information availability and
final-decision quality examined in
\cref{sec:main-results,tab:conditional-outcomes}. In the
first example, the returned information covers 35\% of the sufficient
information set with which it has the greatest overlap. Hence
$D_w(\tau)=0.35$ and $C_w(\tau)=0$, and the trajectory ends with $y=\bot$.
This statement is relative to $\mathcal S(w)$: as established in
\cref{app:acquisition},
$C_w(\tau)=0$ does not rule out a different way of determining the optimum.
In the infeasible example, $C_w(\tau)=1$ and entity grounding is 100\%, yet the
submitted decision has $F_w(y)=E_w(y)=0$. In the feasible but suboptimal
example, $C_w(\tau)=F_w(y)=1$ and $E_w(y)=0$; its normalized regret is 0.507
and its reference utility is 49.3. The examples show why discovery coverage,
entity grounding, feasibility, and objective quality are reported separately.
They illustrate selected information and final-decision gaps rather than all
outcome categories or their prevalence.

\end{document}